\documentclass[acmlarge]{acmart}

\AtBeginDocument{%
  \providecommand\BibTeX{{%
    \normalfont B\kern-0.5em{\scshape i\kern-0.25em b}\kern-0.8em\TeX}}}

\setcopyright{acmcopyright}
\copyrightyear{2023}
\acmYear{2023}
\acmDOI{10.1145/3557885}

\acmJournal{CSUR}
\acmVolume{55}
\acmNumber{9}
\acmArticle{179}
\acmMonth{1}

\usepackage{xcolor}
\usepackage{CJKutf8}
\usepackage[T1]{fontenc}
\usepackage[utf8]{inputenc}
\usepackage{tipa}
\usepackage{newunicodechar}
\newunicodechar{ˈ}{"}
\newunicodechar{ə}{@}
\newunicodechar{ɛ}{E}
\newunicodechar{ɪ}{I}

\begin{document}

\title{Lexical Complexity Prediction: An Overview}


\author{Kai North}
\email{kn1473@rit.edu}
\affiliation{%
  \institution{George Mason University}
  \country{USA}
}
\author{Marcos Zampieri}
\affiliation{%
  \institution{George Mason University}
  \country{USA}
}
\author{Matthew Shardlow}
\affiliation{%
  \institution{Manchester Metropolitan University}
  \country{UK}
}


\renewcommand{\shortauthors}{Kai North, Marcos Zampieri, and Matthew Shardlow.}

\begin{abstract}
    The occurrence of unknown words in texts significantly hinders reading comprehension. To improve accessibility for specific target populations, computational modelling has been applied to identify complex words in texts and substitute them for simpler alternatives. In this paper, we present an overview of computational approaches to lexical complexity prediction focusing on the work carried out on English data. We survey relevant approaches to this problem which include traditional machine learning classifiers (e.g. SVMs, logistic regression) and deep neural networks as well as a variety of features, such as those inspired by literature in psycholinguistics as well as word frequency, word length, and many others. Furthermore, we introduce readers to past competitions and available datasets created on this topic. Finally, we include brief sections on applications of lexical complexity prediction, such as readability and text simplification, together with related studies on languages other than English.
\end{abstract}

\ccsdesc[500]{Reference works~Surveys and overviews}

\keywords{Complex Word Identification, Lexical Complexity Prediction, NLP, Lexical Simplification, Text Simplification, Assistive Technologies.}

\maketitle
\section{Introduction}
\label{intro}

Understanding the meaning of words in context is fundamental for reading comprehension. The perceived difficulty, hereafter referred to as \textit{complexity}, of a target word within a given text varies widely among readers. With an increased demand for distance learning and educational technologies \cite{Morrisetal2020}, research into automatically predicting which words are likely to cause comprehension problems is becoming a popular area of research \cite{paetzold-specia:2016:SemEval1, yiman-EtAl:2018:BEA, semeval-2021}. Systems have been created to identify complex words \textcolor{black}{that are difficult to acquire, reproduce, or understand for} children \cite{kajiwara2013selecting}, second-language learners \cite{lee-yeung-2018-personalizing}, people suffering from a reading disability, such as dyslexia \cite{Rello2013} or aphasia \cite{devlin98, Carroll1998}, or more generally, individuals with low literacy \cite{Watanabeetal2009, Gasperin_etal2009}.

In Computational Linguistics and Natural Language Processing (NLP), the task of automatically recognizing complex words is most often achieved by training machine learning (ML) models. These ML models assign a complexity value to each target word within an \textcolor{black}{inputted extract, sentence, or text} that allows for the identification of complex words. \textcolor{black}{This information can then be used to improve downstream lexical and text simplification systems that provide simpler alternatives to aid reading comprehension. Take the extract shown in Table \ref{example1} for example.}

\begin{table}[!ht]
\centering
\scalebox{1}{
\begin{tabular}{llccccc}
\hline
     Original: & \textbf{Folly} & is & set & in & great & dignity\\
     Simplified: & \textbf{Foolishness} & is & set & in & great & dignity\\
     \hline
\end{tabular}
}
 \caption{\label{example1} Example extract with an identified complex word \textcolor{black}{from} the CompLex dataset \cite{shardlow-etal-2020-complex}.\textcolor{black}{The complex word and its simplified version are in bold.}}
\end{table}

\textcolor{black}{An ML model trained to identify complex words would recognize the word “\emph{folly}” within the original extract as being complex. Such models would come to this decision based on a number of engineered or inferred features. For instance, these models would likely consider the word “\emph{folly}” as being archaic, as having a low frequency within everyday speech, as being unfamiliar to its target demographic or to a general populace, as being acquired later during adolescence, and so forth.} \textcolor{black}{Having identified “\emph{folly}” as being complex, these models may then \textcolor{black}{pass} this information downstream so that it can be simplified to “\emph{foolishness}” resulting in the simplification shown in Table \ref{example1}. Readers may then use this simplification to better understand the meaning of the original sentence or the target word. Alternatively, the simplification of “\emph{folly}” to “\emph{foolishness}” may serve to improve machine translation, since \emph{foolishness} is likely to have a more synonymous equivalent in a target language than “\emph{folly}” \cite{Stajner2016b}. Another use case of identifying complex words is for authorship identification, whereby identifying the number of complex words within a text can serve as a means of measuring vocabulary richness which has traditionally been used as a linguistic fingerprint, hence authorship marker \cite{abdallah2013}.}


\textcolor{black}{The task of identifying complex words is} commonly referred to as Complex Word Identification (CWI) \cite{paetzold-specia:2016:SemEval1}. In recent years, CWI has been extended to Lexical Complexity Prediction (LCP) \cite{shardlow-etal-2020-complex, shardlow2021predicting}\footnote{In this paper, we will be using LCP as the overarching term and CWI specifically when we refer to the binary task of complexity prediction (Section \ref{types_LCP}) \cite{shardlow-etal-2020-complex, shardlow2021predicting}.}. This survey introduces the reader to LCP by providing a comprehensive overview of LCP literature, with a particular focus on the work carried out in the last 10 years \textcolor{black}{that has primarily dealt with English; however, research investigating other languages has also been included and their contributions acknowledged\footnote{\textcolor{black}{We hope that this paper helps to encourage the ongoing development of LCP systems for other languages (Section \ref{other_languages})}}.}


This survey comes at a time of unprecedented demand for LCP research motivated by recent developments in education technology and accessibility, such as the widespread use of virtual learning platforms in distance learning \cite{Morrisetal2020}. It also comes at a time of diversification, with LCP interacting with other topics in NLP, such as machine translation \cite{Stajner2016b} and authorship identification \cite{abdallah2013, srinivasan2019}. To the authors' knowledge, this survey fills a gap in the current LCP literature. It provides new researchers, as well as those who are already familiar with the field, with the most up-to-date key references, \textcolor{black}{main research questions}, advancements, and baselines needed to develop LCP further.

\textcolor{black}{This survey has the following structure. Section \ref{complexity} gives prior definitions of complexity and explains what complexity is, what difficulty is in relation to complexity, and what is meant by the term \textit{complex} in LCP literature. Section \ref{ranking_complexity} briefly describes the origin of complexity prediction research within lexical simplification.} Section \ref{types_LCP} outlines the different types of lexical complexity prediction, ranging from comparative, binary, continuous, and personalized to predicting the complexity of multi-word \textcolor{black}{and numerical expressions.} \textcolor{black}{It also discusses whether systems designed for a target demographic outperform those for a generic population as well as whether predicting the lexical complexity of multi-word expressions is advantageous for LCP (Sections \ref{RQ1} and \ref{RQ2}).} {Section \ref{evaluation_metrics} presents the evaluation metrics used to measure the performance of LCP systems, such as accuracy, precision, recall, F1-score, G-score, mean absolute error, mean squared error, Pearson's correlation, and Spearman's rank.} Section \ref{shared-tasks} details the international competitions that challenged participating teams with the development of LCP systems: CWI--2016 \cite{paetzold-specia:2016:SemEval1}, CWI--2018 \cite{yiman-EtAl:2018:BEA}, ALexS--2020 \cite{Zambrano2020OverviewOA}, and LCP--2021 \cite{semeval-2021}. Section \ref{system_reports} provides a historical overview of the models used for LCP, including feature engineering approaches, neural networks to state-of-the-art transformer-based models. \textcolor{black}{It also describes the best linguistic features for predicting lexical complexity along with the effect including context has on LCP systems' performance (Sections \ref{RQ3} and \ref{RQ4}).} Section \ref{applications} demonstrates LCP's place within \textcolor{black}{the text simplification pipeline and several of its use cases and applications}. Section \ref{resources}  gives an overview of the English datasets and resources used for LCP together with several studies that have investigated languages other than English. \textcolor{black}{It also discusses whether transfer learning is possible for \textcolor{black}{predicting lexical complexity} across multiple languages (Section \ref{RQ6})}. Section \ref{summary} ends by briefly outlining the future of LCP research, including its future opportunities and challenges.




\section{Defining Complexity}\label{complexity}

\textcolor{black}{Within Linguistics, there exists two approaches to defining complexity, when being used to describe the "\textit{complexity}" of a target word: (1) absolute, and (2) relative.} 

\subsection{Absolute Complexity}

\textcolor{black}{Absolute complexity, otherwise known as objective complexity \cite{dahl2004, pallotti2015}, refers to a form of complexity that is established by the objective linguistic properties of a word \cite{bulte_housen2012, pallotti2015}. These linguistic properties include morpho-syntactic, semantic, as well as phonological factors that make a word appear to be complicated, advanced, or convoluted in comparison to a simpler alternative. For instance, having a high number of morphemes, the presence of derivational or inflectional affixes, having multiple meanings, or having multiple vowels or diphthongs, are all characteristic of absolute complexity \cite{ortega2003, bulte_housen2012, pallotti2015}.} 


\begin{equation}\label{absolute_complexity}
  \textbf{un--believ--able} \ or \ \textbf{eng\underline{a}g--ed} \\
\end{equation}

\textcolor{black}{The words, "\textit{unbelievable}" and "\textit{engaged}" both contain two or more morphemes. The word "\textit{engaged}" also has the diphthong \textit{\textipa{eI}} within its first morpheme: \textipa{/ɛnˈg\underline{eɪ}\textdyoghlig/} which \textcolor{black}{is known} to cause production errors for language learners \cite{mousa2015, setyaningsih2019}. "\textit{Engaged}" also has multiple meanings, with one referring to the act of being involved in an activity, and another being pledged to be married \cite{webster_dic}. When used in ambiguous contexts, polysemous words can be troublesome as they hinder a sentence's readability with an example being: "\textit{Do you know if he is available as I think he is engaged?}" \cite{logan_kieffer2017}. Words, such as "\textit{unbelievable}" and "\textit{engaged}", are therefore words with a high degree of absolute complexity since their linguistic properties make them hard to reproduce or understand.} 

\subsection{Relative Complexity}
\textcolor{black}{Relative complexity, also know as agent-related complexity \cite{dahl2004, pallotti2015} or simply referred to as "\textit{difficulty}" \cite{bulte_housen2012}, refers to a type of complexity that is informed by the individual experience or psycholinguistic factors of the individual. For instance, experiences such as the cognitive load, or demand, acquisition difficulty, along with an individual's level of familiarity associated with a particular word or typography, may determine a word's level of relative complexity \cite{ortega2003, bulte_housen2012, pallotti2015}.}

\begin{equation}\label{relative_complexity}
  \textbf{LIKEABLE} \ or \ \textbf{gothic} \\
\end{equation}

\textcolor{black}{\citet{chen_xiao2019} and \citet{liu2022} show that capitalized words are hard for Chinese learners of English to decipher and therefore are cognitively demanding. This is since they have less variance in their overall shape as well as less variance between the shape and size of their individual letters in comparison to words presented entirely in lowercase or Chinese characters which differ greatly in their form:} LIKEABLE versus likeable or \begin{CJK*}{UTF8}{gbsn}讨人喜欢\end{CJK*} (Mandarin for likeable or popular)\textcolor{black}{. Words in reference to a particular art, culture, pop-culture, or historical group are also hard for second-language learners to acquire, especially if no cognate or similar cultural knowledge is available in their native language \cite{stagich1995, yu2020}. Words, such as "\textit{LIKEABLE}" and "\textit{gothic}", are subsequently words with a high degree of relative complexity as factors more associated with the individual, such as typographical unfamiliarity or lack of cultural knowledge, make these words hard to decipher.}

\subsection{Complexity in LCP}

\textcolor{black}{Within LCP research, a more generalized notion of complexity is used. In most cases, the term "{\em complex}" is simply used as a “synonym for difficulty" \cite{malmasi-zampieri:2016:SemEval} and is specifically applied to the word-level, hereby referred to as lexical complexity or complexity. In this field of research, complexity therefore refers to the difficulty an individual may have in acquiring, understanding, or reproducing a particular target word which is often a result of a target word's linguistic properties as well as factors belonging to the individual. Take the following words for example:}

\begin{equation}\label{LCP_complexity}
  \textbf{unbelievable} \ or \ \textbf{gothic} \\
\end{equation}

\textcolor{black}{Both "\textit{unbelievable}" and "\textit{gothic}" have been rated as having a neutral to high degree of lexical complexity within LCP research, regardless of the type of complexity they exhibit, be it either relative, absolute, or both \cite{maddela2018word}. As such, LCP adopts defining characteristics from absolute and relative complexity as determining a word's generalized level of complexity. This generalized notion of complexity is used throughout this paper when referring to the prediction of lexical complexity.} 

\section{Origin of Complexity Prediction}\label{ranking_complexity}

\textcolor{black}{Predicting the lexical complexity of a target word} originated as a sub-task of lexical simplification (LS) \cite{specia2012}. LS aims to replace complex words and expressions with simpler alternatives whilst maintaining the meaning of the original text \textcolor{black}{as exemplified within Table \ref{example1}} \cite{paetzold-specia:2016:SemEval1}. \textcolor{black}{To achieve this}, LCP is used by a LS system for two purposes: (1) to identify complex words that are in need of simplification, and (2) to rank the suitability of simpler alternatives.

\begin{table}[!ht]
\centering
\scalebox{1}{
\begin{tabular}{c|cc}
\hline
     \textbf{Target Word} & \textbf{Rank} & \textbf{Candidate Replacement}  \\
\hline
    Folly & \#1 & Foolishness  \\
    \textbf{} & \#2 & Recklessness  \\
    \textbf{} & \#3 & Silliness  \\
    \textbf{} & \#4 & Craziness   \\
    \textbf{} & \#5 & Stupidity   \\
     
     \hline
\end{tabular}
}
 \caption{\label{tabissuecwi} \textcolor{black}{Examples of candidate replacements generated by an LS system.}} 
\end{table}

\citet{devlin98} and \citet{Carroll1998} were the first to adopt an LCP precursor within their LS systems' pipelines. They used WordNet \cite{miller1995wordnet} \textcolor{black}{as well as} Kučera-Francis's frequency norms, calculated using the Oxford Psycholinguistic Database \cite{wilson1988mrca}, to rank their synonymous and simplified word candidates on what they believed to be their level of complexity. By doing so, their systems provided the most appropriate simplifications for their target complex words \textcolor{black}{allowing for the creation of easier to read texts for aphasic readers; LCP's place within the text simplification (TS) pipeline is described in greater detail within Section \ref{TS_pipeline}}. 

LS--2012 \cite{specia2012} is arguably the first shared-task that contained an LCP element. It tasked five participating teams to design systems to “rank a set of [candidate] words, from the simplest to the most difficult” \cite{Sinha2012}. Participating teams took into consideration a variety of features to conduct complexity prediction. The most common of these features being simplified word candidates' frequency \cite{ligozat-etal-2012-annlor, amoia-romanelli-2012-sb, Sinha2012}, n-grams \cite{ligozat-etal-2012-annlor, Sinha2012}, morpho-syntactic characteristics including context \cite{amoia-romanelli-2012-sb, jauhar-specia-2012-uow}, and psycholinguistic properties \cite{jauhar-specia-2012-uow}.

\section{Types of Complexity Prediction}\label{types_LCP}

\subsection{Comparative Complexity}\label{comparative_complexity}

Using LCP to rank words in terms of their complexity gives rise to a unique type of complexity prediction: comparative complexity. \textcolor{black}{This type of complexity prediction provides a value that is used to distinguish whether a target word is more or less complex than another target word}. As a result, comparative complexity prediction is most often found as a sub-task of LS, rather than its own stand-alone task \cite{Sinha2012, specia2012}. \textcolor{black}{For instance, several studies \cite{jauhar-specia-2012-uow, paetzold-specia-2017-lexical, billami-etal-2018-resyf, north_etal_2022}  have trained various models at comparative complexity prediction with aim of improving LS.} 

\textcolor{black}{\citet{gooding2019_comparative} investigated the effect that comparative judgement labelling had on inter-annotator agreement. They discovered that annotators tasked with ranking the complexity of several target words presented in context agreed more consistently on their chosen labels than compared with annotators tasked with purely identifying complex words without ranking. With a higher rate of inter-annotator agreement comes a higher quality of complexity label, since the true complexity of a target word is more likely to be captured.} \textcolor{black}{As a result, systems trained on such data, or that likewise make comparative judgements, can be highly effective at distinguishing between complex and non-complex words.}


\subsection{Binary Complexity}
\label{CWI}

From 2012 to 2018, complexity prediction research primarily focused on binary complexity prediction. Binary complexity prediction is what is referred to as complex word identification (CWI). CWI is the task of assigning a target word with a binary complexity value of either 1, marking that word as complex, or 0, denoting that word as non-complex.

\begin{table}[!ht]
\centering
\scalebox{1}{
\begin{tabular}{lcccccc}
\hline

     Extract: & \textbf{Folly} & is & set & in & great & \textbf{dignity}\\
\hline
     Complexity Value:  & 1 & 0 & 0 & 0 & 0 & 0\\
     \hline
\end{tabular}
}
 \caption{\textcolor{black}{Example of a sentence annotated with binary complexity values by a CWI classifier taken from the CompLex dataset \cite{shardlow-etal-2020-complex}. Target words of interest are in bold.}}
\end{table}

\noindent \textcolor{black}{CWI is therefore unlike comparative complexity prediction as it purely identifies complex words rather than making comparative judgements or ranking the complexity of simplified word candidates.}

\citet{shardlow2013comparison} \textcolor{black}{was the first to treat CWI as a standalone task separate from LS}. He experimented with a support vector machine (SVM) for CWI and detailed the construction of a binary CWI dataset (Section \ref{CWIcorpus}) together with the impact several features had on his CWI system's performance (Section \ref{system_reports}).

CWI--2016 \cite{paetzold-specia:2016:SemEval1} was the first shared-task that challenged teams directly with binary CWI. This shared-task increased the popularity of complexity prediction research (Section \ref{CWI2016}). However, CWI's modeling as a binary classification task presented a few shortcomings during CWI--2016 \cite{paetzold-specia:2016:SemEval1}. The most notable is that CWI systems were unable to accurately and consistently classify target words on the decision boundary, being those words with an uncertain and often debated level of complexity  \cite{zampieri-EtAl:2017:NLPTEA}.  

\begin{table}[!ht]
\centering
\scalebox{1}{
\begin{tabular}{l|ccccccc|ccc}
\hline
\multicolumn{1}{}{} & \multicolumn{7}{c}{\textbf{Annotations}} & \multicolumn{3}{}{} \\
\hline
     \textbf{Target Word} & \textbf{A} & \textbf{B} & \textbf{C} & \textbf{D} & \textbf{E} & \textbf{F} & \textbf{G} & Avg. & BC & BC Label \\
\hline
     frontier  & \textbf{3} & 4 & 5 & \textbf{3} & 4 & \textbf{3} & \textbf{3}  & 3.57 & 1 & Complex\\
     Milwaukee & 3 & 4 & 4 & 5 & 3 & 2 & N/A & 3.5 & 0 or 1 & Unknown\\
     
     \hline
\end{tabular}
}
 \caption{\label{tabissuecwi} \textcolor{black}{Example of annotator disagreement of the complexity of two target words annotated using a 6-point likert scale. Annotations (or labels) ranged from very simple (1), moderately simple (2), simple (3), to complex (4), moderately complex (5) or very complex (6) with the issue being the even distribution between simple and complex labels. Target words, annotations, and complexity values were taken from the Word Complexity Lexicon \cite{maddela2018word}. \textbf{BC} refers to binary complexity value. Annotations of interest are in bold.}} 
\end{table}

\subsubsection{\bf{Issue with Binary Complexity}}\label{issue_CWI}

Studies have demonstrated that since lexical complexity is subjective and dependent on an individual's experience and a-priori knowledge,  binary CWI is prone to low inter-annotator agreement \cite{maddela2018word, zampieri-EtAl:2017:NLPTEA}. \textcolor{black}{Annotators from different demographics, such as first language or region, have different opinions of what classifies as a complex word, with perceived complexity often changing on an individual-to-individual basis} \cite{maddela2018word, zampieri-EtAl:2017:NLPTEA}. It is this disagreement in whether a word is either a complex or a non-complex word during the annotation process, that creates target words with an uncertain level of complexity \textcolor{black}{that degrades CWI performance}.

\textcolor{black}{Take the words "\emph{frontier}" and "\emph{Milwaukee}" displayed in Table \ref{tabissuecwi} as an example. Within the Word Complexity Lexicon \cite{maddela2018word} that used a 6-point likert scale ranging from very simple (1) to very complex (6), "\emph{frontier}" was given non-complex labels by 4 annotators and complex labels by 3 annotators, whereas "\emph{Milwaukee}" was labeled with non-complex and complex labels 3 times each respectively. Averaging these words' labels we are left with average complexity values that depict an uncertain, or neutral level of complexity given their proximity to the median threshold of 3.5. Converting these annotations to binary complexity values is therefore problematic. The target word "\emph{frontier}" would no longer be considered as being neutral, but rather as being complex as its average complexity value is now over that of the median threshold. This is regardless of the fact that the majority of the labels assigned to "\emph{frontier}" are non-complex. The target word "\emph{Milwaukee}", on the other hand, is a word on the decision boundary, meaning that it can be either labeled as non-complex or complex by a CWI classifier even though its typography may be evidently complex to those whom are unfamiliar with North American loanwords or proper nouns \footnote{It is important to mention that \citet{maddela2018word} may have attempted to avoid such neutral labeling by recruiting an uneven number of annotators, being 11 in total. However, not all annotators labeled each instance.}. Being trained on such examples that have been potentially mislabeled results in CWI systems misclassifying unseen target words. For instance, features used to distinguish non-complex words may be inevitably associated with complex words or vice-versa}. This, in turn, hinders overall CWI performance \cite{zampieri-EtAl:2017:NLPTEA, shardlow-etal-2020-complex}.

\subsection{Continuous Complexity}\label{continuous_complexity}



LCP was introduced to deal with \textcolor{black}{target words with an uncertain level of complexity} along with target words on the decision boundary \cite{maddela2018word, shardlow-etal-2020-complex}. Unlike CWI, LCP alternatively provides a continuous complexity value that is not used to assign a binary complex or non-complex label. Instead, LCP models complexity on a continuum with varying degrees of difficulty \textcolor{black}{with which it then attempts to predict. For instance,} it assigns target words with a complexity label ranging from very easy to very hard that are linked directly to certain thresholds: very easy (0), easy (0.25), neutral (0.5), difficult (0.75), or very difficult (1).

\begin{table}[!ht]
\centering
\scalebox{1}{
\begin{tabular}{ccccccc}
\hline

     Extract: & \textbf{Folly} & is & set & in & great & \textbf{dignity}\\
\hline
     BC & 1 & 0 & 0 & 0 & 0 & 0 \\
     CC & 0.57 & 0 & 0.18 & 0 & 0.15 & 0.42 \\
     \hline
\end{tabular}
}
 \caption{\label{example3}\textcolor{black}{Example of a sentence annotated with binary complexity values (\textbf{BC}), and continuous complexity values (\textbf{CC}) by a LCP regressor taken from the CompLex dataset \cite{shardlow-etal-2020-complex}. Target words of interest are in bold.}}
\end{table}

By modeling complexity on a continuum, LCP provides a more fine-grained representation of the complexity of a target word as it allows for the prediction of more than two levels of difficulty \cite{shardlow-etal-2020-complex}. \textcolor{black}{For example, the target word "\emph{folly}" can be more accurately predicted as a neutral to difficult complex word, whereas the target word "\emph{dignity}" is no longer incorrectly classified as being entirely non-complex (Table \ref{example3})}. An LCP system is thus \textcolor{black}{a linear regressor rather than a binary classifier and for this reason can classify target words that were problematic for prior CWI systems, including such words as "\emph{frontier}" and "\emph{Milwaukee}".}


It is worth mentioning that LCP was not the first to predict lexical complexity as a continuous value. Probabilistic complexity prediction was also a regression based task. However, different from LCP, probabilistic complexity prediction used continuous complexity values to make binary predictions \cite{yiman-EtAl:2018:BEA}. This meant that its continuous complexity values were not used to predict varying degrees of complexity like LCP, but rather to indicate the probability of that target word being either complex or non-complex. CWI--2018 \cite{yiman-EtAl:2018:BEA} developed systems for binary as well as probabilistic complexity prediction. It was the first shared-task that moved away from binary CWI and subsequently laid the foundations for what is known as LCP. CWI--2018 \cite{yiman-EtAl:2018:BEA} is described in more detail within Section \ref{CWI2018}. 


\subsection{Personalized Complexity}\label{personalized_complexity}


Complexity prediction researchers have also been interested in personalizing lexical simplification \cite{tack_etal_2016, lee-yeung-2018-personalizing}. \textcolor{black}{\citet{lee-yeung-2018-personalizing} argued that prior LCP systems are unable to account for “variations in vocabulary knowledge among their users” \cite{lee-yeung-2018-personalizing}}, including other forms of idiosyncrasies, such as cross-linguistic influence \footnote{Cross-linguistic influence being defined as the effects a bilingual speaker's first language (L1) has on their second language (L2) production, hence complexity assignment \cite{yangetal2017}}. \textcolor{black}{Several other researchers \cite{Zengetal_2005, tack_etal_2016, lee-yeung-2018-personalizing} have also} suggested that the previous “one size-fits-all” approach to LCP fails to accurately model varying perceptions of lexical complexity and as a result, personalized CWI was introduced \cite{Zengetal_2005, tack_etal_2016, lee-yeung-2018-personalizing}. This approach creates personalized CWI systems that cater for the individual user or a specific target demographic. These systems are engineered with, or are built to learn, user demographic features that they use to make predictions on an individual basis. These demographic features may include language proficiency, “native language, race, job, age, ethnicity, or education” \cite{Zengetal_2005, tack_etal_2016, lee-yeung-2018-personalizing}.


\subsubsection{{Is personalized complexity prediction worthwhile?}}\label{RQ1}

Personalized complexity prediction systems have been found to outperform LCP systems designed for a generic population when tasked with  predicting the lexical complexities of a target demographic. \textcolor{black}{\citet{Zengetal_2005} discover that demographic features, such as native language, race, job, and so on, can improve CWI performance when predicting the complexity of medical terminology. \citet{tack_etal_2016} created a system designed to predict how well learners of French understood the meaning of a French word by incrementally training their system on features representative of their user's lexical competency. \citet{lee-yeung-2018-personalizing} and \citet{Tack2021} have since implemented personalized CWI models trained on language proficiency and native language \cite{lee-yeung-2018-personalizing, Tack2021}. Both studies demonstrated their personalized CWI systems as outperforming their non-personalized baseline models. \citet{Tack2021} also included contextual features and found that her combined personalized and contextual model outperformed other models that did not take demographic or contextual features into consideration. Personalized complexity is therefore a promising area of complexity prediction research as it is seen to outperform more generalized approaches. Further details regarding a personalized LS dataset are presented in Section \ref{PersonalizedCWIDataset}}.

\subsection{Multi-word Expressions}\label{MWEs_complexity}

LCP as well as other types of complexity prediction are not restricted to predicting the complexity values of single words. Multi-word expressions (MWEs) have also been studied and their complexity values predicted \cite{yiman-EtAl:2018:BEA, semeval-2021}. However, there exists little research into the complexity prediction of MWEs. 

\subsubsection{{Is predicting the lexical complexity of multi-word expressions advantageous?}}\label{RQ2}

According to \citet{goodingetal2020}, assigning complexity values to both single words and MWEs would undeniably improve the performance of LCP systems and, as a consequence, the performance of other downstream NLP-related tasks, such as LS. \citet{goodingetal2020} provide “\emph{ballot stuffing}” as an example. For instance, if complexity values were assigned individually to “\emph{ballot}” and then to “\emph{stuffing}”, this MWE would either not be simplified, as individually  “\emph{ballot}” and  “\emph{stuffing}” may not be considered to be complex words, or simplified into an expression that would be “nonsensical or semantically different” \cite{goodingetal2020}, such as “\emph{ballot filling}” or “\emph{vote stuffing}”. Another example can be seen in Table \ref{exampleMWE}.

\begin{table}[!ht]
\centering
\scalebox{1}{
\begin{tabular}{llcccl}
\hline
     Original: & \textbf{Folly} & is & set & in & great dignity\\
     Simplified: & \textbf{Foolishness} & is & set & in & great dignity\\
     MWE Simplified: & \textbf{Foolishness} & is & set & in & \textbf{pride}\\
     \hline
\end{tabular}
}
 \caption{\label{exampleMWE}\textcolor{black}{Example extract with an identified and simplified MWE taken from the CompLex dataset \cite{shardlow-etal-2020-complex}. Complex words are in bold.}}
\end{table}

\textcolor{black}{As shown in Table \ref{exampleMWE} , “\emph{great}” and “\emph{dignity}”, when taken into consideration separately are not considered to be complex words. However, if these two words were presented to an annotator as one MWE, they may subsequently have been assigned a higher combined complexity value resulting in them as being identified as complex. In turn, a LS system may then provide a more appropriate simplification, such as “\emph{pride}”, that would  further improve the readability of this extract.} For this reason, the CompLex dataset \cite{shardlow-etal-2020-complex} provides 1800 MWEs with preassigned complexity values. \textcolor{black}{LCP--2021 \cite{semeval-2021}} was the first shared-task that challenged teams to develop LCP systems to predict the complexity values of single words and MWEs as two separate sub-tasks. LCP--2021 \cite{semeval-2021} is described further within Section \ref{LCP2021}.


\subsection{Numerical Complexity}\label{numerical_complexity}

\textcolor{black}{Complexity prediction research has also included the identification and simplification of complex numerical expressions. Complex numerical expressions refer to “dates, measurements, quantities, percentages, or ratios” \cite{bautista_saggion_2014}, that children, individuals with poor numeracy, or a learning disability may find to be difficult to interpret \cite{bautista_etal_2011a, bautista_etal_2011b, bautista_saggion_2014}. These numerical expressions can be presented either numerically, for instance,  “\textit{25\%}”, “\textit{>25}” or “$\frac{1}{4}$”, or lexically, as is the case for “\textit{twenty five percent}”, “\textit{greater than 25}”, or “\textit{a quarter}” \cite{bautista_saggion_2014}. The purpose of numerical complexity prediction is to identify which numerical expressions are considered complex and are therefore in need of simplification for a specific target demographic.}

\textcolor{black}{\citet{Rello_etal_2013} conducted an eye-tracking study to gauge the cognitive load associated with numerical expressions when presented as digits compared to when presented as lexical items. They discovered that digits were easier to read for people with dyslexia than compared to words describing numerical expressions.}

\textcolor{black}{\citet{bautista_saggion_2014} have since created a rule-based system for automatically identifying and simplifying complex numerical expressions in Spanish. They hand-crafted numerous rules that utilized regular expressions to identify and then simplify complex numerical expressions within 59 sentences. Their system achieved an F1-score of 0.93 on a manually annotated gold-standard dataset and was subsequently considered to have an acceptable level of performance. \citet{bautista_etal_2017} later incorporated this system within a more generic TS model.}

\section{Evaluation Metrics}\label{evaluation_metrics}

\textcolor{black}{The performance of complexity prediction systems is measured using a variety of evaluation metrics. These evaluation metrics depend on the task, with the most common tasks being: (1) binary classification performed by prior CWI systems \cite{specia2012, paetzold-specia:2016:SemEval1,yiman-EtAl:2018:BEA}, or (2) regression conducted by LCP systems \cite{semeval-2021}, as described within Section \ref{types_LCP}. The following evaluation metrics were used in the international competitions listed within Section \ref{shared-tasks}}.

\subsection{Evaluating CWI Systems}

\textcolor{black}{The performance of binary CWI systems was normally measured using accuracy, precision, recall, F1-score, and G-score. Accuracy is simply the fraction of positive predictions made over the total number observations within the dataset, precision is “the fraction of positive predictions made that are correct” \cite{hackeling2014}, whereas recall is “the fraction of the truly positive instances that the classifier recognizes” \cite{hackeling2014}.}





\paragraph{\bf{F1-Score}} \textcolor{black}{F1-score is the harmonic average of the accuracy and recall scores \cite{hackeling2014}. It is subsequently far more informative for evaluating CWI performance as it penalizes those systems that demonstrate either low precision and recall or a high imbalance between the two \cite{hackeling2014}. Per class F1-scores are then used to calculate macro and weighted F1-scores for all systems. \textcolor{black}{Macro F1-score being the arithmetic mean of all per-class F1-scores, and weighted F1-score being the mean of all per-class F1-scores whilst taking into consideration the number of actual occurrences of each class within the dataset} \footnote{\textcolor{black}{See \citet{hackeling2014} for further details regarding macro and weighted F1-scores as well as how to calculate accuracy, precision, and recall.}}. F1-score is calculated using the equation below (Equation \ref{F1-score})}. 

\begin{equation}\label{F1-score}
F1 = 2 \frac{Precision \cdot Recall}{Precision + Recall}
\end{equation}


\noindent Finally, in CWI 2016 \cite{paetzold-specia:2016:SemEval1}, the organizers used G-scores which, unlike F1-score, takes into account accuracy and recall rather than precision and recall. 


\subsection{Evaluating LCP Systems}
\textcolor{black}{Recent LCP systems designed to predict continuous instead of binary complexity values are commonly evaluated using mean absolute error, mean squared error, Pearson Correlation, and Spearman's Rank.} 

\paragraph{\bf{Mean Absolute Error}} \textcolor{black}{Mean absolute error (MAE) is the average absolute difference between the predicted observations and the actual observations made. It is calculated using the following equation (Equation \ref{mae}).}

\begin{equation}\label{mae}
    MAE = \frac{\sum_{i=1}^{n}|y_i - x_i|}{n}
\end{equation}

\noindent \textcolor{black}{where \textit{n} is the total number of observations, \textit{i} is the current observation, \textit{y} is the predicted observation, and \textit{x} is the actual observation seen. The closer a MAE value is to zero, the greater the system's performance.}

\paragraph{\bf{Mean Squared Error}} \textcolor{black}{Mean squared error (MSE) is the average squared difference between the predicted observations and the actual observations made. MSE is used to understand the variance and the bias of the predicted observations. Variance refers to the spread of the predicated observations. Bias refers to spread of the predicted observations compared to that of the actual observations. MSE is produced by the following equation (Equation \ref{mse}).}

\begin{equation}\label{mse}
    MSE = \frac{\sum_{i=1}^{n}(y_i - x_i)^2}{n}
\end{equation}  

\noindent \textcolor{black}{where \textit{n} is once again the total number of observations, \textit{i} is the current observation, \textit{y} is the predicted observation, and \textit{x} is the actual observation seen. An MSE closer to zero may indicate the presence of less outliers within the provided dataset.}

\paragraph{\bf{Pearson's Correlation}} \textcolor{black}{Pearson's Correlation (R) was the primary means of evaluation in LCP--2021 (See Section \ref{LCP2021}) \cite{semeval-2021}. It measures the linear relationship between two variables and returns a value between -1 and 1 with a returned value closer to -1 indicating a strong negative correlation, whereas a returned value closer to 1 indicating a strong positive correlation. Pearson's correlation is calculated using the following equation (Equation \ref{pcc}).}

\begin{equation}\label{pcc}
    R_{X,Y} = \frac{cov(X, Y)}{\sigma_X\sigma_Y}
\end{equation}

\noindent \textcolor{black}{where \textit{X} and \textit{Y} are the variables taken into consideration, $\sigma$ is the standard deviation and \textit{cov(X,Y)} stands for co-variance of the two variables.}

\paragraph{\bf{Spearman's Rank}} \textcolor{black}{Spearman's Rank ($\rho$) takes into consideration the monotonic relationship between two variables, even if this relationship is not linear. Therefore, unlike Pearson's Correlation, Spearman's Rank is more robust when dealing with outliers. It also returns a value between -1 and 1 that depicts the same associated correlations: strong negative (-1), strong positive (1). Spearman's Rank is provided by the follow equation (Equation \ref{rho}).}

\begin{equation}\label{rho}
    \rho = 1- {\frac {6 \sum d_i^2}{n(n^2 - 1)}}.
\end{equation}

\noindent \textcolor{black}{where \textit{$d_i$} is the difference between the two ranks of each observation, and \textit{n} is the total number of observations.}

\section{International Competitions}\label{shared-tasks}

LCP has been the focus of several international competitions, known as shared--tasks. These shared--tasks have been described throughout the following sections (Sections \ref{CWI2016} to \ref{LCP2021}). Further detail regarding the architecture, development, and evolution of the systems submitted to these shared--tasks has been provided in Section \ref{system_reports}. In addition, system summaries have been provided in the Appendices in Tables \ref{table_CWI2016} to \ref{table_LCP2021}.


\subsection{CWI--2016 at SemEval}\label{CWI2016}

The first CWI shared-task, referred to as CWI--2016, was organized at the International Workshop on Semantic Evaluation (SemEval).\footnote{\url{http://alt.qcri.org/semeval2016/task11/}.} CWI--2016 (SemEval--2016 Task 11) was modelled as a binary classification task. Participants developed systems to predict the complexity value of English words in context. The organizers provided a dataset sampled from various sources such as the CW Corpus \cite{shardlow2013comparison}, the LexMTurk Corpus \cite{Horn2014}, and Simple Wikipedia \cite{coster-kauchak-2011-simple}. 

The target words in the CWI--2016 dataset were annotated by a pool of 400 non-native English speaking annotators. The CWI--2016 dataset was split into a training and a test set. The training set included 2,237 target words in 200 sentences each annotated by 20 annotators. A word was considered complex in the training set if at least one of the 20 annotators assigned it as such. The test set included 88,221 target words in 9,000 sentences each annotated by a single annotator. According to the organizers of CWI--2016, this setup was devised to imitate a scenario where the goal was to predict the individual needs of a speaker based on the needs of the target group  \cite{paetzold-specia:2016:SemEval1}. Finally, in terms of task setup, CWI--2016 considered only single word annotations while MWEs were not considered.


A total of 21 teams submitted 42 systems to CWI--2016 and 19 of them wrote system description papers published in the SemEval proceedings. Participants used a wide range of models and features summarized in Table \ref{table_CWI2016} (Appendices) and discussed further in Section \ref{system_reports}.  

Most teams who participated in the shared-task used simple probabilistic models trained on features such as n-grams, word frequency, and word length. The approaches used by the top-3 systems in CWI--2016, being PLUJAGH \cite{wrobel:2016:SemEval}, LTG \cite{malmasi-dras-zampieri:2016:SemEval}, and MAZA \cite{malmasi-zampieri:2016:SemEval}, also relied on probabilistic classifiers and on the aforementioned features. The F1-scores achieved by the top-3 systems were 0.353, 0.312, and 0.308 respectively which were considered rather low compared to the baselines and the post-competition analysis presented in \citet{zampieri-EtAl:2017:NLPTEA}. According to \citet{zampieri-EtAl:2017:NLPTEA}, this indicated that CWI--2016 was a particularly challenging task due to the data annotation protocol and the training/test split, since 40 times more testing data was available compared to the training data.

\subsection{CWI--2018 at BEA}\label{CWI2018}

The second edition of the CWI shared-task,\footnote{\url{https://sites.google.com/view/cwisharedtask2018/}.} referred to as CWI--2018, was organized at the Workshop on the Innovative Use of NLP for Building Educational Applications (BEA). CWI--2018 was a multilingual shared-task featuring datasets containing English, French, German, and Spanish data. A total of three tracks were available, namely English, German, and Spanish monolingual, with a fourth additional track being made available at a later date. Furthermore, training and testing data from the multi-domain \emph{CWIG3G2} dataset \cite{yimam-EtAl:2017:RANLP} was available for each initial language. The fourth track was the French multilingual track where only a French test set was available and the participants had to use the data made available for the other three languages to make predictions in French (Section \ref{cross-lingual_LCP})\footnote{See Section \ref{cross-lingual_LCP} for more information regarding this type of complexity prediction: cross-lingual LCP.}. 

The CWI--2018 datasets were split on training, development and testing partitions. The English dataset contained 27,299 instances for training, 3,328 for development, and 4,252 for testing. The Spanish dataset featured 13,750 instances for training, 1,622 for development, and 2,233 for testing. The German dataset included 6,151 for training, 795 for development, and 959 for testing. Finally, the French dataset only included a testing partition with 2,251 instances.

The three main new aspects of CWI--2018 compared to CWI--2016 were: (1) its multilingual nature compared to the English-only CWI--2016, (2) the presence of both target single words and multiple consequent words, and (3) two sub-tasks, one modelled as a binary classification task, and one modelled as a probabilistic classification task.

CWI--2018 received submissions by 12 teams in multiple task and track combinations. At the end of the competition, 10 teams wrote system description papers presented at the BEA workshop. In Table \ref{tab:approaches2018} (Appendices), we present the approaches by teams who submitted systems to the CWI--2018 English binary classification task and who also wrote system description papers. An observed trend was that more teams tried deep neural networks in CWI--2018 compared to CWI--2016, a trend also observed in other areas of AI and NLP research (Section \ref{system_reports}). 

In CWI--2018's binary classification task, being sub-task 1, the organizers reported the performance from all teams in each of the three domains, namely \emph{News}, \emph{WikiNews}, and \emph{Wikipedia}. As discussed in the CWI--2018 report \cite{yiman-EtAl:2018:BEA}, the performance obtained by all teams on the News domain was generally substantially higher than the performance obtained in the two other domains.



\subsection{ALexS--2020 at SEPLN}\label{ALexS2020}

\textcolor{black}{ALexS--2020 \cite{Zambrano2020OverviewOA}, referring to the lexical analysis shared-task at the Intentional Conference of the Spanish Society on Natural Language Processing (SEPLN), was the first shared-task to look at CWI for Spanish educational texts.}  

\textcolor{black}{The shared-task included a Spanish dataset consisting of 9,175 words, with 723 of these words being identified by 430 student annotators as complex. These words were taken from transcripts of academic videos in Spanish made within the University of Guayaquil, Ecuador. Teams were challenged with creating a system to automatically identify which of these 9,175 were labeled as complex.} 


\begin{table*}[!ht]
\centering
\scalebox{0.76}{
  \begin{tabular}{lp{5cm}p{9.8cm}c}
\hline
  \bf Team & \bf Classifiers & \bf Features & \bf Paper \\ \hline

\bf   UDLAP & Threshold-based & General lexicon, specialized lexicon of internet-related terms, n-grams, frequency. & \cite{Sulayes2020GeneralLC} \\

\bf   Vicomtech & Gaussian Mixture Models (GMM) and K-Means Clustering & Lemma length, lemma frequency in subject documents, number of synsets in WordNet, lemma frequency in domain corpora, lemma probability in domain corpora, word frequency in Wikipedia and word probability in Wikipedia. & \cite{Zotova2020VicomtechAA} \\

\bf   HULAT & Support Vector Machine & Word length, a boolean determining whether only capital letters were used, a boolean determining a target words inclusion in an easy-to-read lexicon, Word2Vec vectors and BERT vectors. & \cite{Alarcn2020HulatA} \\

  \hline
  \end{tabular}
}
\caption{Systems submitted to the ALexS--2020 in alphabetical order as summarized by \cite{Zambrano2020OverviewOA}.}
\label{tab_ALexS2020}
\end{table*}

\textcolor{black}{Three teams participated at ALexS--2020. Each team was presented with the entire dataset, with only the total number of complex words being revealed. As such, no development or training partitions were provided. This encouraged the development of several models as shown within Table \ref{tab_ALexS2020}.} 

\textcolor{black}{The performances achieved at ALexS--2020 were considered to be poor. The best performing system by UDLAP \cite{Sulayes2020GeneralLC} attained a macro F1-score of 0.272, whereas the best performing systems of VIcomtech \cite{Zotova2020VicomtechAA} and HULAT \cite{Alarcn2020HulatA} achieved macro F1-scores of 0.176 and 0.164 respectively. These low performances indicated the overall difficulty of the task, since not being presented with a training or development set lead to the teams having no idea what was considered to be characteristic of a complex word within the particular domain of Spanish educational texts.}

\subsection{LCP--2021 at SemEval}\label{LCP2021}

The 2021 Lexical Complexity Prediction Task \cite{semeval-2021}, referred to as LCP--2021, was also held at SemEval and attracted 58 teams across its 2 sub-tasks as shown within Table \ref{table_LCP2021} (Appendices). 

The dataset \cite{shardlow-etal-2020-complex} was developed using crowd sourcing. 10,800 instances were selected from three corpora covering the Bible \cite{Christodouloupoulos2015}, biomedical articles \cite{koehn2005europarl} and europarl \cite{bada2012concept}. LCP--2021's dataset contained single words (9,000 instances) and MWEs (1,800 instances). The MWEs were limited to pairs of nouns, or adjective-noun collocations. The annotated tokens were presented in context to both the original annotators and the participating teams. This meant that the complexity assignments were not only for the token, but instead for the token in its contextual usage. Multiple instances of tokens were included in different contexts, each receiving differing contextual complexity assignments. As such, systems that took context into account fared well in the final evaluation.


The organizers split the dataset into trial, train and test sets, stratifying the data for the token type, token instance, complexity and genre. This meant that even distributions of MWEs and single words were available in each subset as well as an even distribution across genres. Complexity labels were also evenly distributed between the subsets with each having a similar spread of labels. The repeated occurrences of tokens were grouped together in each subset, such that no subset shared any tokens with another subset to prevent information bleed between subsets.

The shared-task allowed participants to submit to one of two sub-tasks. The first sub-task permitted systems to only predict the complexity values of the single word instances within the CompLex dataset \cite{shardlow-etal-2020-complex}. The second sub-task asked participants to predict the complexity values for the entire dataset, forcing them to develop a methodology for adapting their single word models to the MWE use case. The organizers did not evaluate solely on MWEs due to the smaller size of the subset. All data was collected via CodaLab and the systems were ranked according to their Pearson's Correlation with the held-back gold standard labels on the test sets.



Several of the top-ranking systems for LCP--2021's sub-task 1 used transformer-based models \cite{vaswani2017attention}. However, systems that used hand-crafted features \cite{semeval2021_task1_paper_62,semeval2021_task1_paper_64,semeval2021_task1_paper_68} also performed well with the top performing system \cite{semeval2021_task1_paper_148} in this category having achieved third place on the official ranking table. This is discussed further within Section \ref{other_SOA}.

Sub-task 2 saw fewer participants than sub-task 1 (37 teams in total). Systems used similar models to those in sub-task 1, with the key difference being the strategy for combining MWEs. Feature-based systems were able to average the features \cite{semeval2021_task1_paper_57,semeval2021_task1_paper_68,semeval2021_task1_paper_99} or predictions \cite{semeval2021_task1_paper_62} for each token in an MWE to give the overall value. Deep learning based systems were typically able to encode the MWE as part of their existing training scheme by supplying the transformer architecture with two encoded tokens instead of one.

\section{Approaches to Predicting Lexical Complexity in English Texts}
\label{system_reports}


Various ML models have been used for LCP. These range from support vector machines (SVMs), decision trees (DTs), random forests (RFs), neural networks to state-of-the-art transformers, such as BERT \cite{devlin2019bert}, RoBERTa \cite{liu2019roberta} and ELECTRA \cite{clarketal2020}. Many of these models have also been used in unison to form ensemble-based models. Prior to more recent transformer-based models, ensemble-based models that utilized multiple DTs, RFs, or neural networks, were state-of-the-art in predicting lexical complexity \cite{paetzold-specia:2016:SemEval1, yiman-EtAl:2018:BEA}. This section describes in detail the various models used for LCP. It demonstrates the evolution of LCP systems by providing their model's architecture and performance.

\subsection{Machine Learning Classifiers}\label{systems2013-2018}

\subsubsection{\bf{Support Vector Machines}}\label{SVMs}

SVMs are statistical classifiers. They use labeled training data and engineered features to predict the class of unseen inputs \cite{Cortes&Vapnik1995_SVM, shardlow2013comparison}. SVMs are well suited for binary classification. They achieve exceptional performance when there exists a clear distinction between two classes. SVMs work less well when dealing with multiple classes or a large number of features as this reduces the uniqueness of each class. SVMs were popular within early LCP research which focused on binary complexity prediction \cite{specia2012, jauhar-specia-2012-uow}. 

\citet{jauhar-specia-2012-uow} were one of the first to adopt an SVM for complexity prediction. They trained their SVM on three types of features: morphological, contextual, and psycholinguistic. Morphological features were generated through the use of character n-grams. Contextual features were obtained through a bag-of-words approach, whereby n-grams were used to select neighbouring words. Psycholinguistic features were in relation to a target word's degree of concreteness, imageability, familiarity, and age-of-acquisition. Their SVM outperformed a prior baseline CWI model trained on word frequencies. 

\citet{shardlow2013comparison} created a complex word corpus (the CW Corpus) consisting of 731 complex words in context \cite{CWcorpus} (Section \ref{CWIcorpus}). He then experimented with a variety of simplification techniques, including the use of a SVM for binary complexity prediction. His SVM was trained on several features. These features being word frequency, syllable count, word senses, and synonyms associated with the target word. His SVM achieved a higher recall over its precision. This indicated that his SVM was good at identifying complex words, yet often missclassified non-complex words as being complex. It was subsequently prone to the word boundary misclassification problem that is associated with binary CWI systems (Section \ref{issue_CWI}). 

\citet{kuru:2016:SemEval} was interested in the use of Glove word embeddings \cite{pennington2014glove} for capturing the contextual information of a target word. Building on \citet{jauhar-specia-2012-uow}'s bag-of-words approach in extracting contextual information, \citet{kuru:2016:SemEval} investigated how effective Glove word-embeddings, or vectors representations, were at CWI when used as features. They trained two SVM models, referred to as AIKU native and AIKU native1, which they submitted to CWI--2016 \cite{paetzold-specia:2016:SemEval1}. The first model: AIKU native, was trained on the “word embedding of the target word and its substrings as features” \cite{kuru:2016:SemEval}. The second model: AIKU native1, was trained on the word embedding of the target word, its substrings, as well as the embeddings of the target word's neighbouring words. They discovered that both of their models performed equally well having attained matching G-scores of 0.545 at CWI--2016 \cite{paetzold-specia:2016:SemEval1}. This led \citet{kuru:2016:SemEval} to conclude that contextual information, such as a target word's neighbouring words, was not a useful feature in improving the CWI performance of a SVM model.

\citet{sp-kumar-kp:2016:SemEval} experimented with Word2vec word embeddings alongside statistical, POS-tag, and similarity features. They trained four SVM models. Their first model was trained on Word2vec word embeddings. Their second model was trained on Word2vec word embeddings, word length, number of syllables, ambiguity count, and frequency. Their third model was trained on Word2vec word embeddings and the similarities between the target word and its neighbouring words. Their fourth model was trained on all of the above features, taking into consideration word embeddings, along with statistical and contextual features. The fourth model was found to be the best. Submitted as AmritaCEN (w2vecSim) to CWI--2016, it achieved a F1-score of 0.109 and a G-score of 0.547 \cite{paetzold-specia:2016:SemEval1, sp-kumar-kp:2016:SemEval}. It comes as no surprise that given their reliance on word-embeddings and contextual information, \citet{sp-kumar-kp:2016:SemEval}'s AmritaCEN (w2vecSim) and \citet{kuru:2016:SemEval}'s AIKU (native1) have both achieved similar performances. However, an interesting observation is that \citet{sp-kumar-kp:2016:SemEval}'s fourth model with the addition of POS-tags: AmritaCEN (w2vecSimPos), performed less well. This would suggest that POS-tags are less important for CWI than previously theorized. This is supported by the performance of POS-tags as a feature for LCP as shown by \citet{LCP-RIT}.

\subsubsection{\bf{Decision Trees}}

DTs make predictions based on a set of learned sequential or hierarchical rules housed in decision nodes, or leafs. They apply a top-down approach, filtering labeled data through various decision nodes, or branches, until that data is separated as accurately as possible in accordance to class. As such, DTs are often found to surpass the performance of SVMs at LCP \cite{paetzold-specia:2016:SemEval1}. This may be due to DTs being better suited in dealing with features that overlap between classes, given their reliance on learned rules rather than prototypical features, such as support vectors. 

Throughout CWI--2016, as detailed in Section \ref{CWI2016}, the most common and arguably the most successful CWI systems consisted of either a DT or a Random Forest (RF) model \cite{paetzold-specia:2016:SemEval1}. This marked LCP's transition to DTs and RFs. These models maintained state-of-the-art status until LCP--2021 \cite{semeval-2021}. This is partly due to these models being trained on a greater number of varied and unique features related to lexical complexity. The use of these additional features was inspired by \citet{shardlow2013comparison}, \citet{jauhar-specia-2012-uow}, and others' success at surpassing previous baseline performances. It is also partly due to the use of DTs and RFs within ensemble-base models; this is described in greater detail within Section \ref{ensembles}. 

\citet{choubey-pateria:2016:SemEval} investigated the performance of both a SVM and a DT at CWI. They discovered that their “SVM seemed to be less effective for CWI” \cite{choubey-pateria:2016:SemEval, paetzold-specia:2016:SemEval1}. Their SVM attained a F1-score of 0.179 and a G-score of 0.508, whereas their DT produced a F1-score of 0.181 and a G-score of 0.529 \cite{choubey-pateria:2016:SemEval}. They reasoned that their SVM's slightly worst performance was due to it having “overlapping decision boundaries” \cite{choubey-pateria:2016:SemEval}. Again, this refers to the decision boundary misclassification problem that is commonly faced by CWI systems (Section \ref{issue_CWI}). 

The systems submitted by \citet{quijada-medero:2016:SemEval}, referred to as team HMC, were among the top performing systems at CWI--2016 \cite{quijada-medero:2016:SemEval, paetzold-specia:2016:SemEval1}. One of HMC's systems consisted of a DT, known as HMC-DecisionTree25, whereas the another consisted of a regression tree (RT), named HMC-RegressionTree05. These models outperformed their SVM counterpart, with the DT model achieving a F1-score of 0.298 and a G-score of 0.765. Both models were set to have a maximum depth of three meaning that only three decision nodes, or rules, were learned. These rules were learned from several inputted features. These features belonged to two main categories: statistical, and psycholinguistic \footnote{HMC also utilized POS-tags as features.}. Their statistical features included unigram and lemma frequencies, word, stem and lemma length, probability of a word's character sequence, and lastly, number of synsets, whereas their psycholinguistic features included age-of-acquisition, perceived word concreteness and the number of differing pronunciations associated with a target word \cite{quijada-medero:2016:SemEval}. They claimed that their models' success was due to their use of corpus-based features, especially their use of unigram and lemma frequencies. 

\subsubsection{\bf{Random Forests}}

RFs consist of multiple DTs. Each DT is trained on a random subset of the training data. From their limited input, each DT then learns a sequence of hierarchical rules for classification. A RF's final output is generated through a plurality voting system. Since each DT only observes a small fraction of the training data, it results in RFs being less prone to overfitting. Each DT learns to distinguish its inputted classes without making sweeping generalizations across the entire dataset. This means that each DT becomes specialized at identifying the distinguishing features of its limited input. Pooling these DTs together subsequently makes for a RF that is more adaptable to unseen data than a stand-alone DT. A RF is, therefore, better suited at dealing with a large dataset with a large number of features compared to a single DT.

\citet{ronzano-EtAl:2016:SemEval} submitted a RF to CWI--2016 that outperformed other DT models \cite{paetzold-specia:2016:SemEval1}. Their RF, referred to as TALN (RandomForest\_WEI), was taken from the Weka machine learning framework \cite{Hall2009}. Being a RF, it consisted of several DTs trained on multiple features, many of which being similar to the features used by the two HMC systems \cite{quijada-medero:2016:SemEval}. However, like other models submitted to CWI--2016, additional features were also exploited, such as contextual features \cite{paetzold-specia:2016:SemEval1}. These contextual features took into consideration the position of the target word within a sentence, the number of tokens within that sentence, and the frequencies of both the target word and its context words within the British National Corpus (BNC) \cite{BNC, leech2014word} and the 2014 English Wikipedia Corpus \cite{WikiCorpus} \footnote{The presence of low or high frequency context words was believed to be an indicator of a target word's degree of complexity. If on average, a target word was surrounded by more highly frequent context words, then that target word was believed to be non-complex, whereas if it were surrounded by less frequent words, then that target word was believed to be complex.}. The use of such contextual features, together with its RF architecture, may explain TALN's superior performance in comparison to HMC's DT and RT models \cite{quijada-medero:2016:SemEval}. TALN (RandomForest\_WEI) achieved an F1-score of 0.268 and a G-score of 0.772. This was respectively -0.02 less than the F1-score and +0.006 better than the G-score achieved by the best performing HMC system \cite{paetzold-specia:2016:SemEval1, ronzano-EtAl:2016:SemEval}.

\citet{zampieri-tan-vangenabith:2016:SemEval} created a CWI system, referred to as MACSAAR (RFC), with a particular focus on Zipfian features. Zipf's Law implies that words that appear less frequently within a text are longer and as a result are likely to be considered more complex than words that are more frequent and shorter \cite{zampieri-tan-vangenabith:2016:SemEval, quijada-medero:2016:SemEval}. To test this assumption, they trained a SVM, RF, and nearest neighbor classifier (NNC) using a variety of Zipfian features. These features included word frequency, word and sentence length, and the sum probabilities of the character trigrams belonging to the target word or to the sentence. Their RF model was their best performing model. It attained a F1-score of 0.270 and a G-score of 0.754 at CWI--2016 \cite{paetzold-specia:2016:SemEval1} giving it a greater F1-score of +0.002, yet an inferior G-score of -0.018 compared to TALN \cite{ronzano-EtAl:2016:SemEval}. Per their model's performance, \citet{zampieri-tan-vangenabith:2016:SemEval} concluded that Zipfian features are good baseline indicators of lexical complexity.

\citet{davoodi-kosseim:2016:SemEval} experimented with several models for CWI--2016 \cite{paetzold-specia:2016:SemEval1}. These models were a naïve bayes, a neural network, a DT, and a RF. Their best performing model was their RF, referred to as CLacEDLK (CLacEDLK-RF\_0.6). This model was trained on several features. \citet{davoodi-kosseim:2016:SemEval} had a particular interest in psycholinguistic features, namely abstractness. They believed there existed a correlation between “the degree of abstractness of a word and its perceived complexity”\footnote{Non-complex words are theorised to have more concrete meanings than complex words, hence complex words are believed to be more abstract in regards to their meaning \cite{Brysbaert2013}.} \cite{davoodi-kosseim:2016:SemEval}. They developed two RF models. Their first RF had a threshold of 0.5, whereas their second RF had a threshold of 0.6. This meant that for a target word to be classified as being complex, these RFs' sub-DTs' output would have on average a complexity value above 0.5 for their first RF and above 0.6 for their second RF. Their second RF was found to outperform their first by a G-score of +0.028. As such, having a higher threshold for complexity assignment would appear to improve CWI performance.

\subsubsection{What are the best linguistic features for predicting lexical complexity?}\label{RQ3}

\textcolor{black}{The SVMs, DTs, and RFs described above have all so far utilized a common set of features that can be separated into four categories: statistical, morpho-syntactic, psycholinguistics, and contextual.  Work by \citet{LCP-RIT}, \citet{Tack2021}, as well as \citet{shardlow2021predicting} have since demonstrated that such statistical features, such as word length, word frequency and syllable count, psycholinguistic features, including prevalence (average familiarity), age-of-acquisition, and concreteness, together with contextual features, the likes of character or word-level n-grams, continue to be good predictors of lexical complexity.} 


\textcolor{black}{Recently, \citet{LCP-RIT} went as far as to rank the effectiveness of several features using a RF trained on the CompLex dataset \cite{shardlow-etal-2020-complex} (Section \ref{LCP2021}). They discovered that prevalence, age-of-acquisition, and concreteness achieved the first, second, and third best performances respectively and POS-tags and prior complexity labels achieved the worst performances. However, apart from the use of character-level bigrams,  \citet{LCP-RIT} failed to investigate the effect contextual features would have had on their model's performance. Contextual features have also been exploited in ensemble-based models, neural networks, and state-of-the-art transformers. The impact of these models' use of contextual features is discussed in Section \ref{RQ4}.}

\subsection{Ensemble-based Models}\label{ensembles}

A RF is an ensemble-based model. An ensemble-based model is any model that is made up of multiple sub-models and that produces a final output through some form of plurality voting. These sub-models can be of the same type, as is the case for an RF, or of differing types. The main advantages of ensemble-based models are brought about through their diversity. An ensemble-base model can utilize the strengths of various models, be it either SVMs, DTs, RFs, neural networks, or even transformers, whilst simultaneously mitigating the disadvantages associated with using only one type of model. As a consequence, ensemble-based models are state-of-the-art for LCP. However, throughout the years, differing combinations of sub-models have been used. From CWI--2016 \cite{paetzold-specia:2016:SemEval1} to CWI--2018 \cite{yiman-EtAl:2018:BEA}, the best performing ensemble-based models consisted of a combination of DTs, RFs, or neural networks. Since LCP--2021, this has changed. State-of-the-art ensemble-based models now consist of various transformers (Section \ref{state-of-art}).

\citet{malmasi-zampieri:2016:SemEval} built upon the use of multiple DTs, hence a RF for binary CWI. They adopted a meta-classifier architecture. A meta-classifier architecture is a unique type of ensemble-based model. It “is generally composed of an ensemble of base classifiers that each make predictions for all of the inputted data” \cite{malmasi-zampieri:2016:SemEval}. These base classifiers then input their output into a second set of classifiers. This second set of classifiers, or meta-classifiers, take as features the output of the first set of base-classifiers. They then produce their own output through “a plurality voting process” \cite{malmasi-zampieri:2016:SemEval}.

\citet{malmasi-zampieri:2016:SemEval} submitted two ensemble-based models to CWI--2016: MAZA A and MAZA B \cite{paetzold-specia:2016:SemEval1}. Both of these models' base classifiers were decision stumps, which are different from DTs as they are trained on a single feature and subsequently only have one decision node, thus giving them the appearance of a tree stump rather than of an entire tree. Bootstrap aggregation was then applied to the output of each decision stump. This bagged output was then inputted into a second level of meta-classifiers consisting of “200 bagged decision trees”\cite{malmasi-zampieri:2016:SemEval}. 


MAZA B was trained using additional contextual features that were not utilized by MAZA A \cite{malmasi-zampieri:2016:SemEval}. These contextual features were also different from those used by other aforementioned systems. Together with word frequencies, MAZA B also incorporated two types of probability scores as contextual features. The first being conditional probabilities, being the probability of a target word appearing next to its neighbouring one or two words. The second being joint probabilities, being the probability of a target word occurring in conjunction with its surrounding words within a sentence. As such, MAZA B was found to outperform MAZA A. It achieved a F1-score of +0.116 greater than MAZA A \cite{malmasi-zampieri:2016:SemEval, paetzold-specia:2016:SemEval1}. \citet{malmasi-zampieri:2016:SemEval} contributed this superior performance to MAZA B's use of contextual features, highlighting the importance to which they believed context influences a word's perceived level of complexity\footnote{\citet{malmasi-zampieri:2016:SemEval} would appear to contradict \citet{kuru:2016:SemEval}, as \citet{kuru:2016:SemEval} found context to be uninfluential on his SVM's performance. Early LCP research debated the importance of context. However, context is now more firmly believed to be an influential factor within current LCP literature \cite{semeval2021_task1_paper_148, north_etal_2022} (See Section \ref{RQ4} for further details).}.

\citet{choubey-pateria:2016:SemEval} constructed two ensemble-based models for CWI--2016 \cite{paetzold-specia:2016:SemEval1}. The first, referred to as GARUDA (HSVM\&DT), had a meta-classifier architecture which comprised of five SVMs and five DTs. In this model, the SVMs were the base classifiers tasked with the binary classification task of CWI. Its second set of meta-classifiers were its DTs. These meta-classifiers identified whether the predictions made by its SVMs were correct or incorrect.  \citet{choubey-pateria:2016:SemEval}'s second ensemble-based model contained twenty SVMs. Unlike their first model, their second model did not employ meta-classifiers. Instead, each of the 20 SVMs were tasked with predicting the labels of the entire training set. The best performing SVMs then had the most impact in calculating the model's final output labels through a performance oriented voting system. Interestingly, their first ensemble-based model was found to perform worst than individual SVM or DT models, whereas their second ensemble-based model achieved average performance. They blamed this poor performance on the “overlapping decision boundaries” \cite{choubey-pateria:2016:SemEval} of their SVM sub-models. This once again demonstrates the inferiority of SVMs for CWI compared to other models.

The SV000gg systems, created by \citet{paetzold-specia:2016:SemEval1}, were the best performing systems submitted to CWI--2016 \cite{paetzold-specia:2016:SemEval1, paetzold-specia:2016:SemEval2}. \citet{paetzold-specia:2016:SemEval1} adopted ensemble-based models that utilized a variety of sub-models. They believed that model diversity would result in greater CWI performance. They experimented with  ensemble-based models that consisted of a lexicon-based model, a threshold-based model to SVMs, DTs, RFs and other machine learning classifiers. Their lexicon-based model identified whether a target word was a complex or a non-complex word by searching for that word within a given dictionary of pre-labeled lexemes. Their threshold-based model separated complex and non-complex words by seeing whether a target word had a particular feature above a certain threshold and that was also found to be a defining characteristic of that word type; see Section \ref{TS_pipeline} for more information regarding lexicon-based and threshold-based approaches to predicting lexical complexity. 

The predictions made by their diverse set of sub-models were counted and then used to determine the system's final output through hard or soft voting. As such, there were two versions of the SV000gg system: Hard SV000gg and Soft SV000gg. Hard SV000gg used hard voting to produce the final output label by counting how many times in total the target word was labeled as being either complex or non-complex by all of its contained sub-models. Soft SV000gg used a form of performance-oriented soft voting. Traditional soft-voting generates a summed confidence estimate in regards to how likely a target word belongs to a particular class. The final label assigned to this word is then resulted from this summed confidence estimate. Performance-orientated soft voting determines the final label of a target word by examining the performances of each sub-model “over a certain validation set such as precision, recall, and accuracy” \cite{paetzold-specia:2016:SemEval2}. The most common label produced by these sub-models with the highest overall performance, is then chosen as the final output label.

Soft SV000gg achieved the best performance with an F1-score of 0.246 and a G-score of 0.774. Hard SV000gg attained a slightly worst F1-score and G-score of 0.235 and 0.773 respectively. However, Hard SV000gg still outperformed all of the other systems submitted to CWI--2016 in regards to its G-score, including those mentioned above \cite{paetzold-specia:2016:SemEval1}. As a result, both models demonstrated the superiority of diverse ensemble-based models for binary CWI in comparison to other models.

\citet{syspaper7} were inspired by the performance of prior ensemble-based models at CWI--2016 \cite{paetzold-specia:2016:SemEval1}. Their system, referred to as Camb, ranked first on both of CWI--2018's sub-tasks: binary CWI and probabilistic complexity prediction, when dealing with English monolingual data (Section \ref{shared-tasks}) \cite{yiman-EtAl:2018:BEA}. Camb used a boosting classifier: AdaBoost, with 5000 estimators followed by a RF bootstrap aggregation model \cite{syspaper7}. They experimented with differing sub-models, each being trained on a set of given features similar to those used by prior CWI systems \cite{paetzold-specia:2016:SemEval1}. They concluded that an ensemble-based model that combines both AdaBoost and a RF with equal weights, consistently produced the best performance \cite{syspaper7}. 

\citet{syspaper11} experimented with the tree learner model provided by KNIME \cite{Bertholdetal_2009}, along with other combinations of DTs, RFs, and gradient boosted tree learners for CWI--2018's sub-task 2: probabilistic complexity prediction \cite{syspaper11, paetzold-specia:2016:SemEval1}. They found that their KNIME tree learner model obtained good results when set to contain 600 models. It achieved a mean macro F1-score of 0.818 across the three datasets provided by CWI--2018 (Section \ref{shared-tasks}). Therefore, \citet{syspaper7} and \citet{syspaper11} have demonstrated that ensemble-based models achieve good performance at binary as well as probabilistic complexity prediction.

\subsection{Neural Networks}\label{neural_networks}

Deep learning is highly popular within NLP and Computational Linguistics having achieved state-of-the-art performance in various NLP-related tasks \cite{Venkatetal2018, wu2021graph}. Neural networks attempt to mimic human learning by artificially replicating the neuroplasticity of the human brain. They achieve this by manipulating weight values (synaptic strength) between nodes (neurons) that contain characteristic information, or learned features, related to the input (or environmental experience as is the case with the human brain). These weight values are adjusted through a loss function applied after each epoch, or iteration. This process is repeated until these weight values are fully optimized and the most optimum output is produced.
 
Neural networks can be either supervised or unsupervised. This means that they can learn such characteristic information, or features associated with a complex word, independently. However, within LCP research, neural networks have consistently under-performed in comparison to other more traditional feature engineered models, such as DTs or RFs. This is especially true when such traditional models have been combined within ensemble-based models \cite{paetzold-specia:2016:SemEval1, yiman-EtAl:2018:BEA}. It was not until the introduction of continuous complexity prediction in the form of probabilistic complexity (Section \ref{continuous_complexity}), that some neural networks were shown to perform well, and on occasion, on par with more traditional models \cite{syspaper11, yiman-EtAl:2018:BEA}.

\citet{nat:2016:SemEval} was one of the first to investigate the performance of a recurrent neural network (RNN) at binary CWI. Within their RNN, they included a gated recurrent unit (GRU). A GRU is designed to safeguard against the vanishing gradient problem. The vanishing gradient problem arises during back-propagation, when the neural network adjusts its loss function in accordance to its current prediction. The vanishing gradient problem refers to when the gradient of the loss becomes excessively small overtime, thus, inhibiting the weight values of earlier nodes from being accurately updated \cite{nat:2016:SemEval, Gillioz2020}. This impairs a neural network's ability to retain information learned at earlier stages. A GRU counters this problem by acting as a “memory” device \cite{nat:2016:SemEval}. It controls what new information should be learned, what prior information should be remembered, and what previous information should be forgotten, when updating a weight value.

\citet{nat:2016:SemEval} created an RNN model with a GRU as well as a ensemble-based model with a meta-classifier architecture. Referred to as Sensible (Combined), their ensemble-based model was built up of five RNNs as base classifiers and a single RF as a meta-classifier. Out of all of the neural network models submitted to CWI--2016, their RNN model with a GRU, referred to as Sensible (baseline), achieved the best performance \cite{nat:2016:SemEval, paetzold-specia:2016:SemEval1}. Nevertheless, in comparison to other more traditional models, Sensible (baseline) performed poorly. It attained an F1-score of 0.140 and a G-score of 0.646. \citet{nat:2016:SemEval} claimed it was the small size of CWI-2016's training set that caused their RNN model to perform less well than expected (Section \ref{CWI2016}). 

\citet{syspaper11} were the first to experiment with a convolutional neural network (CNN) for binary CWI. A CNN is different from a RNN. It contains an additional convolutional layer that takes as input the output of its first layer and then transforms said input before passing it onto a further layer. However, CNN models lack the temporal capabilities of an RNN with an embedded GRU. Regardless of this limitation, the CNN introduced by \citet{syspaper11}, referred to as NLP-CIC-CNN, slightly outperformed their ensemble-based model, consisting of various KNIME tree learners, on one out of the three datasets provided by CWI--2018 \cite{yiman-EtAl:2018:BEA}  (Section \ref{ensembles}). It attained a macro F1-score of 0.855 and an accuracy rating 0.863. This surpassed the macro F1-score and accuracy achieved by their ensemble-based model by +0.003 and +0.004 respectively. 

\citet{syspaper4} compared models that adopted feature engineering to neural networks at CWI--2018 \cite{yiman-EtAl:2018:BEA}. They trained a variety of models, such as DTs, Gradient Boosting, Extra Trees, AdaBoost and XGBoost methods, on numerous features including statistical features, such as word length, number of syllables, numbers of senses, hypernyms and hyponyms, along with n-gram log probabilities; again, being similar to those features previously used by prior CWI systems (See Tables \ref{table_CWI2016} \& \ref{tab:approaches2018}). These models were compared to a shallow neural network that used word embeddings, and a Long Short-Term Memory (LSTM) language model capable of handling the vanishing gradient problem through its use of a forget gate along with a additive gradient structure; being parallel to the use of a GRU. 

For binary CWI \cite{yiman-EtAl:2018:BEA}, \citet{syspaper4}'s feature engineered XGBoost model outperformed their neural network models. It attained an F1-score of 0.8606, whereas their shallow neural network and LSTM models achieved F1-scores of 0.8467 and 0.8173 respectively. Nevertheless, for CWI--2018's second sub-task of probabilistic complexity prediction, their LSTM model, referred to as NILC, was superior to all of the other models, having achieved a F1-score of 0.588. Their feature engineered XGBoost model, and their shallow neural network model, achieved less impressive F1-scores of 0.2978 and 0.2958 respectively. Both \citet{syspaper11} and \citet{syspaper4}, therefore, proved the viability of using neural networks for probabilistic complexity prediction.

\subsubsection{\bf{Transformers}}
\label{state-of-art}

The best performing systems of LCP--2021 \cite{semeval-2021} used transformer-based models. Transformer-based models were introduced to overcome the limitations associated with prior neural networks, such as RNNs, and LSTM models \cite{nat:2016:SemEval, syspaper4, syspaper11}. \textcolor{black}{ \citet{vaswani2017attention} outlines several advantages of transformers, namely their self-attention mechanism and their ability to more effectively capture long-term dependencies. }




Just Blue by \citet{semeval2021_task1_paper_148}, achieved the highest Pearson's Correlation at LCP--2021's sub-task 1 of 0.7886 \cite{semeval-2021}. It was inspired by the prior state-of-the-art performance of ensemble-based models together with the recent headway in various NLP-related tasks made by transformers \cite{semeval2021_task1_paper_148}. 

Just Blue consisted of an ensemble of BERT \cite{devlin2019bert} and RoBERTa \cite{liu2019roberta} transformers. This system contained two BERT models as well as two RoBERTa models. Bert1 and RoBERTa1 were fed target words, whereas Bert2 and RoBERTa2 were fed the target words' corresponding sentences, hence context. These models then predicted the lexical complexities of their inputted target words or sentences, whereby their outputted complexity values were determined by weighted averaging. Models 1 had a weight of 80\% and models 2 had a weight of 20\%. This meant that the complexity of target words was considered to be more important than the complexity of their surrounding words. Howbeit, each sentence was still taken into consideration when calculating the weighted average, as prior studies have shown context to be an influential factor on continuous complexity prediction \cite{malmasi-zampieri:2016:SemEval, quijada-medero:2016:SemEval}. Once a weighted average was returned by either set of models: BERT and RoBERTa, Just Blue's final output was produced as a simple average of these returned weighted averages. 


\citet{semeval2021_task1_paper_148} experimented with different models as well as different weight splits between their target word and sentence level inputs. They discovered that between SVM, RF, BERT, and RoBERTa models, along with a BERT and RoBERTa hybrid model, a BERT and RoBERTa hybrid model achieved the highest performance. They also found that between a 90/10, 80/20, and a 70/30 split between target word and sentence level input, a 80/20 weight split, being in favor of the target word, produced the most accurate complexity values. As such, Just Blue's success is likely a result of its diverse ensemble of varying models, as well as its use of, but not over-reliance on, a target word's context.

 DeepBlueAI developed by \citet{semeval2021_task1_paper_74} achieved second place at LCP--2021's sub-task 1 and first place at sub-task 2 \cite{semeval2021_task1_paper_74, semeval-2021}. It attained a Pearson's Correlation of 0.7882 for sub-task 1 and a Pearson's Correlation of 0.8612 for sub-task 2. It used a variety of pre-trained language models, such as the transformers BERT \cite{devlin2019bert}, RoBERTa \cite{liu2019roberta}, ALBERT \cite{lan2020albert}, and ERNIE \cite{sun2019ernie}. DeepBlueAI was subsequently an ensemble-based model that used model stacking with five layers. All of its aforementioned transformers were utilized within its first layer. Its second layer then adjusted the transformers' hyperparameters. It manipulated dropout, the number of hidden layers, and the loss function. The third layer then conducted 7-fold cross-validation to check for overfitting or selection bias, with the fourth layer then having adopted training strategies, such as data augmentation and pseudo-labelling. Data augmentation is the training strategy of adding new data to a training set by copying and slightly modifying existing data; in this instance, data from CWI--2018 was used, and for sub-task 2, data from sub-task 1 was used after having gone through “synonym replacement, random insertion, random swap, and random deletion” \cite{Wei&Zou2019, semeval2021_task1_paper_74}. Pseudo-labelling is the training strategy of predicting labels for unlabeled data and then adding the newly labeled data back into the training set. The fifth layer contained DeepBLueAI's final estimator in the form of a simple linear regression model. This estimator returned the final predicted complexity values (${\hat{y}}$) through the following equation (Equation \ref{DeepBlueAI}):

\begin{equation}\label{DeepBlueAI}
\hat{y} = \sum_{j=1}^{N}W_{j}\hat{y}_{j}
\end{equation}

\noindent where \textit{${N}$} is the total number of transformers with different hyperparameters, \textit{${W_{j}}$} is the weight of each transformer, and \textit{${\hat{y}_{j}}$} is each transformers' predicted complexity value.

\citet{semeval2021_task1_paper_74} attributed their model's good performance in both sub-tasks to its use of multiple transformers and training strategies. With model diversity also being an influential factor in regards to Just Blue's high performance \cite{semeval2021_task1_paper_148}, it would appear that current state-of-the-art LCP systems consist of an ensemble of differing transformers-based models.

RG\_PA, created by \citet{semeval2021_task1_paper_107}, was the second highest performing system at LCP--2021's sub-task 2 having achieved a Pearson's Correlation of 0.8575 \cite{semeval2021_task1_paper_107, semeval-2021}. Unlike Just Blue \cite{semeval2021_task1_paper_148} and DeepBlueAI \cite{semeval2021_task1_paper_74}, it did not contain an ensemble of diverse transformers. Alternatively, RG\_PA consisted of a single RoBERTa attention based model. It used Byte-Pair Encoding (BPE) to firstly tokenize all of its inputted sentences. BPE compresses a given sentence so that its most frequent character pairs, or bytes, are replaced with a single character. This shortens the inputted sentence into a sequence of character representations that help to mitigate the out-of-vocabulary problem \footnote{The out-of-vocabulary problem refers to the problem that arises when a model is presented with a word that was not observed within its training set.}. Each of their RoBERTa's hidden layers applied token pooling that creates a vector representation of a target word based on the average of all of the token embeddings of that target word found throughout the training set. The attention weight between the target vector and context tokens, i.e. context words, is then calculated and the returned context vector is concatenated with the target vector. The concatenated vector representation of each target word is then used to predict the complexity values of the unseen words within the test set. Its use of BPE together with its use of concatenated context and target word vectors, may explain RG\_PA's high performance in sub-task 2, despite it not being an ensemble-based model.

\subsection{Other State-of-the-Art Models}\label{other_SOA}

The third best performing system at LCP--2021's sub-task 1, deviated from the use of transformer-based models \cite{semeval2021_task1_paper_62}. \citet{semeval2021_task1_paper_62} approached sub-task 1 from a more traditional feature engineering approach. Much like prior CWI sytems, \citet{semeval2021_task1_paper_62} utilized a combinations of lexical, contextual, and semantic features (Section \ref{CWI}). However, unlike previous CWI systems, these features were extensive with 51 features in total being used to rate lexical complexity. These features included SUBTLEX features, word etymology, and several readability indices. SUBTLEX features are those features that are embedded within film subtitles, such as the number of films whose subtitles depict the word in lowercase, target word frequency per million subtitled words, as well as the percentage of films where the target word appeared within the SUBTLEX-US corpus \cite{brysbaert2009moving}. Features related to a word's etymology included the number of Greek or Latin affixes that belong to the target word, and readability index features included Flesch score \cite{Flesch1948}, Gunning-Fog index \cite{Gunning1952}, LIX score \cite{Anderson1983}, SMOG index \cite{McLaughlin1969}, and Dale-Chall index \cite{Chall-Dale1995}.

\citet{semeval2021_task1_paper_62} fed his extensive list of features into a Light Gradient Boosting Machine (LGB) model with minimal optimization. Results showed that the top three most influential features on LCP performance were age, the Dale-Chall index, and the complexity values taken from \citet{maddela2018word}'s word complexity lexicon. \citet{semeval2021_task1_paper_62} also observed that several sentence readability features were top contributors with the Dale-Chall index being the most influential. The Dale-Chall index is a readability index that measures a text's perceived comprehension difficulty by assessing its words familiarity based on a list of 3,000 common words \cite{Chall-Dale1995}. It is likely that his extensive list of features, the inclusion of such contextual features, or contextual readability measures, along with his use of a LGB model, is responsible for \citet{semeval2021_task1_paper_62} outperforming a similar feature engineering approach by \citet{LCP-RIT}.

\subsubsection{What effect does the inclusion of context have on predicting lexical complexity?}\label{RQ4}
\textcolor{black}{Several prior LCP studies that have not yet been mentioned have demonstrated mixed results when it comes to the inclusion of context. On the one hand, \citet{syspaper2} experimented with three configurations of the dataset provided by CWI--2018 \cite{yiman-EtAl:2018:BEA}: “context-free, context-only, and context-sensitive" \cite{syspaper2} and found no significant difference in their systems performance between the three. Furthermore, \citet{krizetal2018} discovered an increase in their CWI system's performance when one neighbouring word was taken into consideration, yet a decrease in its performance when this was increased to two or more neighbouring words. On the other hand, \citet{gooding-kochmar-2019-complex} improved their winning system at CWI-2018 \cite{yiman-EtAl:2018:BEA}: SV000gg, by capturing contextual information as a result of converting their model to a sequence labelling task and by using word-embeddings. In addition, several studies throughout Sections \ref{SVMs} to \ref{neural_networks}, have claimed that the inclusion or inference of contextual features is responsible for their systems high performance \cite{sp-kumar-kp:2016:SemEval, kuru:2016:SemEval, malmasi-zampieri:2016:SemEval}. The transformer-based models outlined in Section \ref{state-of-art} also make no exception to this claim and due to their superiority over prior neural networks, such as their ability to more effectively capture long-term dependencies, their exploitation of contextual information is likely even more beneficial \cite{semeval2021_task1_paper_148, semeval2021_task1_paper_148}. In the past, the effect context had on LCP performance was therefore somewhat debated. However, the high performance of recent models that have included sentence and word-level features, now more definitely suggest that the inclusion of context improves LCP.}



\subsection{Summary}\label{models_summary}

Current state-of-the-art LCP systems consist of an ensemble of varying transformers. These systems achieve state-of-the-art performance largely due to two reasons: (1) \citet{semeval2021_task1_paper_148} and \citet{semeval2021_task1_paper_74} have demonstrated the importance of model diversity within an ensemble-based model, and (2) transformer-based models are better equipped to handle contextual information (Section \ref{neural_networks}). As such, ensembles-based models that rely on multiple transformers of varying types and that take into consideration contextual information of the target word, are currently the state-of-the-art systems for LCP. However, \citet{semeval2021_task1_paper_62} has proven that feature engineering is still a viable approach for LCP, given that an extensive set of lexical, contextual, and semantic features are taken into consideration. 

\section{Use cases and applications}\label{applications}

LCP has many potential use cases and applications \cite{paetzold-specia:2016:SemEval1, yiman-EtAl:2018:BEA, semeval-2021}. LCP systems can be utilized within a variety of assistive technologies, such as computer-assisted language learning (CALL) applications or intelligent tutoring systems (ITSs) to improve the readability of given texts (Section \ref{readability}). This is most often achieved by implementing text simplification (TS) that benefits from a LCP component.

\subsection{Improving Readability}
\label{readability}

CALL is the use of any computer related technology, be it either a word processing document, social media, or other online medium, for language learning. ITSs are “computer learning environments designed to help students master difficult knowledge and skills” \cite{Gresseretal2018}. CALL applications subsequently include ITSs that specialize in language learning and have been found to improve second language (L2) acquisition \cite{Tseng&Yeh}. These applications include multiple designs, are based on differing pedagogical practices, and allow for varying degrees of learner-computer interaction \cite{alkhatlan2018intelligent}.


A common approach among CALL is to simplify a text to make it more accessible for the L2 learner \cite{Rets&Rogaten2020, Zamanetel2020}. \citet{Alhawiti2015} states that TS can be beneficial to language learners and therefore, an ITS or CALL application that incorporated TS would likewise be beneficial. This is since TS has been found to increase the literacy \cite{petersen2009} as well as advance the vocabulary development of L2 learners \cite{tack-etal-2016-evaluating, Rets&Rogaten2020}.


\citet{Rets&Rogaten2020} tested 37 participants on their ability to memorize and process the ideas presented within two texts: (1) an authentic text, and (2) a simplified text with less complex vocabulary and syntax. Memory was measured by asking the participants to rewrite the observed texts, whereas text processing was gauged through the use of eye tracking. Participants were found to achieve greater memorization and were shown to fixate less on the simplified text than compared to the authentic text. This led \citet{Rets&Rogaten2020}) to conclude that TS results in better textual comprehension which correlates with a greater learning potential \cite{Pulido2004, Rets&Rogaten2020}.

ITSs that use TS are not restricted to aiding L2 learners. TS improves the readability of texts and thus enhances the literacy development of other target demographics. TS may help an ITS designed for people diagnosed with autism “by reducing the amount of figurative expressions in a text” \cite{sikka2020survey}. It may also increase the effectiveness of ITSs created for people with dyslexia or aphasia. This is by replacing long words with short words, or substituting words with challenging character combinations for those which are easier to identify \cite{Rello2013, Carroll1998}. ITSs developed for children may likewise use TS in order to reduce the amount of high-level jargon, or uncommon words, within a text \cite{DeBelder&Moens2010}. TS is, therefore, useful in improving the vocabulary and literacy development of L2 learners \cite{Alhawiti2015}, people with autism \cite{sikka2020survey}, dyslexia \cite{Rello2013, Carroll1998}, or aphasia \cite{Carroll1998}, as well as children \cite{DeBelder&Moens2010}. \textcolor{black}{However, \citet{crossley_2007} presents arguments for and against the use of simplified texts within L2 classrooms with \citet{gooding-2022-ethical} pointing out that the usefulness of simplified texts may vary between target demographics and in some instances may be inferior to alternative reading strategies}.  \textcolor{black}{Despite this, throughout the years TS systems have assessed lexical complexity through a number of ways.}


\subsection{LCP's Place in the Text Simplification Pipeline}
\label{TS_pipeline}


\textcolor{black}{Prior to the LCP systems outlined within Section \ref{system_reports}, TS assessed lexical complexity through several approaches: (1) a simplify everything approach, (2) a threshold-based approach, and (3) a lexicon-based approach \cite{Paetzold_Specia2017}. \textcolor{black}{However, each of these approaches had their limitations that led to the development of more dynamic LCP systems.}}

\subsubsection{\bf{Simplify Everything}}
\label{simplify_everything}

\textcolor{black}{The simplify everything approach simplified all of the words within a given text \cite{devlin98}. This approach subsequently had no means of identifying complex words. Instead, systems that adopted this approach often used a form of comparative complexity prediction to compare and find the most suitable word replacements for every single word within a provided text. A disadvantage to this approach is that not all words are in need of simplification \cite{shardlow14a_ls}. As such, systems that adopted this approach often simplified already easy to understand words into equally easy to understand alternatives that were not as well suited as the original word for that particular context \cite{devlin98, paetzold-specia-2013-text}. The simplify everything approach was therefore found to produce ungrammatical and nonsensical simplifications.} 

\subsubsection{\bf{Threshold-Based}}
\label{threshold_based}

\textcolor{black}{Threshold-based approaches required the presence of a feature over a set value in order for a target word to be identified as complex. Systems that adopted this approach often used a single feature-threshold, such as having \textit{x} number of characters, or \textit{x} frequency in a certain corpus, as a means of gauging the complexity of a target word \cite{Paetzold_Specia2017}. However, this approach was found to be insufficient in identifying all instances of complex words within a given text. For instance, \citet{bott-etal-2012-spanish} and \citet{Shardlow2014} discovered that using word length as a standalone feature-threshold failed to identify complex words which were uncharacteristically short, whilst incorrectly classifying simple words that were over 5 characters long. An example being incorrectly classifying \textit{folly} as non-complex yet \textit{foolishness} as complex, since the former may be considered a short word and the latter a long word. As such, the reliance of a single, or sometimes multiple, feature-thresholds lost popularity as an accurate means of assessing lexical complexity.}  

\subsubsection{\bf{Lexicon-Based}}
\label{lexicon_based}

\textcolor{black}{Lexicon-based approaches utilized a predefined list of words as a means of distinguishing between complex and non-complex words within a given text \cite{Paetzold_Specia2017}. Systems that adopt lexicon-based approaches are often found to perform well in identifying complex words for their intended target demographic or domain. However, when identifying complex words for individuals outside of their intended target population or domain, lexicon-based approaches perform less well \cite{Paetzold_Specia2017}. For example,  FACILITA \cite{Watanabe2010C} is designed to distinguish between Portuguese complex and non-complex words for Brazilian children using three dictionaries: (1) consisted of frequent words extracted from Brazilian newspapers, (2) contained concrete words, and (3) housed simple words that were believed to be “common to youngsters” \cite{Aluisio2010}. FACILITA is very effective in helping young low literacy readers in Brazil. However, FACILITA may be less helpful for other demographics, such as second language learners, individuals suffering from a reading disability, or older individuals with low literacy. This is since the words used to make FACILITA's predefined dictionaries may not be considered as easy to understand for these demographics as they were for Brazilian children.}

\subsection{Other Use Cases}\label{use_cases}

LCP can aid other downstream NLP-related tasks, such as machine translation \cite{Stajner2016b} and authorship identification \cite{abdallah2013, srinivasan2019}, and is also likely to be beneficial to other downstream tasks within the the future. Two alternative use cases of LCP are exemplified in the following sections (\ref{MT} to \ref{AI}).


\subsubsection{\bf{Machine Translation}}\label{MT}
Before TS shifted to improving the readability of texts, its primary focus was to aid machine translation (MT) \cite{Al-Thanyyan_etal_2021}. MT is the task of automatically translating a source language into a target language \cite{Stajner2016b}. MT systems are limited by the lack of parallel corpora that contain identical texts in more than one language. MT systems are also hindered by the morpho-syntactic complexities of the languages that they are tasked to translate. Studies have proven that TS can aid MT \cite{Stajner2016b, stajner-popovic-2019-automated, Al-Thanyyan_etal_2021}. TS achieves this by reducing the ambiguity of the inputted texts in the source language \cite{Stajner2016b}. For instance, by replacing complex words in the source language with simpler alternatives, it increases the probability of an MT system finding a suitable translation in the target language.

\begin{table}[!ht]
\centering
\scalebox{1}{
\begin{tabular}{cc}
\hline

      & Sentence\\
\hline
     Original &  A dozen Chinese fishing \textbf{boats} had taken \textbf{refuge} in a lagoon of Huangyan Island \\
     Simplified & A dozen Chinese fishing \textbf{ships} had taken \textbf{shelter} in a lagoon of Huangyan Island \\
     \hline
\end{tabular}
}
 \caption{\label{TS_MT_example}\textcolor{black}{Example of a simplified sentence shown in \citet{Stajner2016b}. Target words of interest are in bold.}}
\end{table}

\textcolor{black}{\citet{Stajner2016b} demonstrated that a TS system that utilized both LS and syntactic simplification components improved the performance of a English-to-Serbian MT system. Their system, being assessed on the adequacy (meaning preservation) and fluency (grammatical correctness) of its output, achieved this by translating simplified sentences rather than translating the original sentences directly.}

\textcolor{black}{According to \citet{Stajner2016b}, the simplified sentence shown in Table \ref{TS_MT_example}, resulted in a English-to-Serbian translation that was both easier to understand and more grammatically correct to a group of Serbian annotators than compared to a translation of the original sentence. Without a LCP component, the simplified words: \textit{boats} and \textit{refuge}, may not have been recognized as being complex and as a consequence, would not have been simplified resulting in a less accurate translation. This demonstrates that the inclusion of an LCP component within the TS pipeline can improve MT.}

\subsubsection{\bf{Authorship Identification}}\label{AI}
Authorship identification is the task of identifying the author of a given text \cite{Boran2020}. A text's vocabulary richness is a common feature used for authorship identification. Vocabulary richness is used to capture an individual's linguistic fingerprint, in other words, their idiolect. It is normally measured through the use of the type-token ratio (TTR). The TTR is "a simple ratio between the number of types and tokens within a text" \cite{kubat2014}. The TTR, therefore, shows the diversity of a author's vocabulary. It has been used in such situations as helping to differentiate between authors of highly similar texts \cite{rexha2018} as well as to identify the author of online messages \cite{srinivasan2019}. 

LCP provides an additional measurement of vocabulary richness. Adding to the TTR, it provides an average lexical complexity marker that depicts, on average, how complex the author writes. Average lexical complexity can be inputted into an authorship identification system as a feature that may, in turn, enhance its performance. \textcolor{black}{\citet{tanguy2012} experimented with such a feature, in the form of morphological (lexical) complexity, for the authorship identification of various extracts taken from fictional books. Alternative examples are using lexical complexity to differentiate between authors belonging to different time-periods, authors with different levels of education, or authors of different ages; under the assumption that  \textcolor{black}{discrepancies exist} between their writing styles. For instance, past literature may contain vocabulary considered to be more archaic and complex than modern literature, individuals with a higher level of education may use more jargon-related and complex words than those with a lower level of education, and adults may use more unfamiliar and less common words than children.}


\section{Resources}\label{resources}

\subsection{Additional English Datasets and Resources}\label{English_datasets}
The shared-tasks of CWI--2016 \cite{paetzold-specia:2016:SemEval1}, CWI--2018 \cite{yiman-EtAl:2018:BEA}, and LCP--2021 \cite{semeval-2021}, tested participating teams on three datasets that have since contributed significantly to LCP research (Section \ref{shared-tasks}). Nevertheless, these are not the only influential datasets that contain words with lexical complexity ratings. All of the current LCP datasets that deal with English and that are known to the authors' are provided in Table \ref{tab:datasets} located within the Appendices. Apart from the CWI--2016, CWI--2018, and the CompLex datasets already discussed within Section \ref{shared-tasks}, the remaining datasets are introduced throughout the following sections (\ref{CWIcorpus} to \ref{PersonalizedCWIDataset}).

\subsubsection{\bf{CW Corpus}}\label{CWIcorpus}
The CW Corpus contains 731 complex words in context \cite{CWcorpus}. It was constructed using wikipedia edits. Edits are often made to Wikipedia entries in order to simplify their vocabulary. Using Wikipedia's edit history, it is possible to see the simplified edit as well as the original text. To determine which of these edits contained true lexical simplifications, \citet{CWcorpus} looked at the editor's comments for the word "simple", and calculated Tf-idf vector representations to check for lexical discrepancies between the original and simplified texts. Those texts which were found to contain true lexical simplifications, were then subject to a set of further tests to guarantee the validity of the CW corpus. Hamming distance was calculated to ensure that only one word differed between the original and simplified texts. Reality and inequality checks were conducted to make sure that the target words were known yet different English words, and not just variations of the same word. Lastly, non-synonym pairs were discarded and simplified candidate words were verified. Through these series of checks, 731 complex words were provided with context.

\subsubsection{\bf{\citet{Horn2014}}}\label{Middlebury Corpus}
\citet{Horn2014} created a corpus of 25,000 simplified word candidates for comparative complexity prediction. They acquired 50 annotators. These annotators were required to live in the US in an attempt to control their English proficiency. They were asked to give a simpler alternative for each target complex word within 500 sentences. They achieved this by using Amazon’s Mechanical Turk (MTurk) that is popular among NLP-related tasks \cite{Horn2014}. Similar to \citet{CWcorpus}, the sentences presented to the annotators were taken from a sentence-aligned Wikipedia corpus. This corpus provided original and simplified Wikipedia entries of the same texts. On average, annotators provided 12 differing simplifications per target word. This makes the corpus introduced by \citet{Horn2014} a valuable resource for investigating comparative complexity.

\subsubsection{\bf{Word Complexity Lexicon}}\label{WCL}
\citet{maddela2018word} recognized the limitations of prior CWI datasets, namely, the limitations associated with using binary complexity labels rather than continuous complexity values (Section \ref{issue_CWI}) \cite{maddela2018word}. As a response to these limitations, they constructed the Word Complexity Lexicon (WCL). The WCL is a dataset made up of “15,000 English words with word complexity values assessed by human annotators” \cite{maddela2018word}. These 15,000 words were the most frequent 15,000 words found within the Google 1T Ngram Corpus \cite{Brants&Franz2006_GoogleNgram_corpus}. Their assigned word complexity values were continuous since these values were assigned by 11 non-native yet fluent English speakers using a six-point likert scale. They assigned each word with a value between 1 and 6, with 1 denoting that word as being very simple, and 6 defining that word as being very complex. To determine the final complexity value of each word, complexity values were averaged. Those complexity values which were greater than 2 from the mean of the rest of the ratings, were discarded from the final average. This improved the WCL's inter-annotator agreement to 0.64. The remaining disagreements between the annotators were believed to be due to the differing characteristics of their native languages, hence caused by cross-linguistic influence. 

\subsubsection{\bf{Personalized LS Dataset}} \label{PersonalizedCWIDataset}
\citet{lee-yeung-2018-personalizing} constructed a dataset of 12,000 words for personalized CWI. These words were ranked on a five-point likert scale. 15 learners of English, who were native Japanese speakers, were tasked with rating the complexity of each of the 12,000 words. The five labels that they could choose from ranged between (1) “never seen the word before”, to (5) “absolutely know the word's meaning” \cite{lee-yeung-2018-personalizing}. \citet{lee-yeung-2018-personalizing} converted these multi-labeled ratings into binary labels. They considered words ranked 1 to 4 as being complex, and words ranked 5 as being non-complex. However, their use of a multi-labeled likert scale means that this dataset can be used for continuous complexity prediction.

The 15 annotators chosen for data annotation were split into two groups of English proficiency. Thus, two subsets of the dataset were created: the low English proficiency subset, and the high English proficiency subset. The low English proficiency subset was annotated by learners whom knew less than 41\% of the 12,000 words. The high English proficiency subset was annotated by learners whom knew more than 75\% of the 12,000 words. As such, the Personalized LS Dataset \cite{lee-yeung-2018-personalizing} is an ideal resource for future personalized LCP research.

\subsection{Lexical Complexity Prediction in Languages Other than English}\label{other_languages}

Since CWI--2018 \cite{yiman-EtAl:2018:BEA} (Section \ref{CWI2018}), LCP for other languages has began to receive more attention in the form of monolingual, multilingual, and cross-lingual LCP \cite{yeung-lee-2018-personalized, finnimore-etal-2019-strong}. Monolingual LCP refers to the the task of predicting the complexity values of words in a single language. Multilingual LCP refers to the task of creating a LCP system that can be trained on and used to predict the lexical complexities of multiple languages. Cross-lingual LCP refers to the task of training a LCP system on one or multiple languages and then using that system to predict the lexical complexities of a language previously unseen within the training set. 


\subsubsection{\bf{French, Spanish and German}}

\textcolor{black}{As previously mentioned in Section \ref{CWI2018}, the CWI--2018 shared-task at BEA \cite{yiman-EtAl:2018:BEA}, contained datasets in French, Spanish, and German. It was discovered that systems generally performed well across these languages, with high performance in one language correlating with high performance in another. The organizers of CWI--2018 saw this as evidence in support of cross-lingual LCP (Section \ref{cross-lingual_LCP}).}


\citet{billami-etal-2018-resyf} was interested in the perceived lexical complexity of French words and as a result created the ReSyf lexicon. \textcolor{black}{This lexicon contains French synonyms that have been ranked in regards to their reading difficulty using a SVM ranker trained on the Manulex resource} \cite{Lete_etal_2004}. \citet{gari-soler-etal-2018-comparative} investigated the performance of word embeddings at predicting the lexical complexity of French words. They discovered that word embeddings outperformed statistical features, such as word length, number of phonemes, or log frequency when used in isolation. However, when these statistical features were used in unison, they outperformed word embeddings. Other studies interested in French, such as \citet{tack_etal_2016} and \citet{Tack2021}, have already been described in the Personalized Complexity Section (\ref{personalized_complexity}).

\textcolor{black}{The ALexS--2020 \cite{Zambrano2020OverviewOA} shared-task and its submitted systems \cite{Sulayes2020GeneralLC, Zotova2020VicomtechAA, Alarcn2020HulatA} sought to predict Spanish lexical complexity and have been introduced within Section \ref{ALexS2020}. \citet{merejildo2021} has since detailed that construction of a Spanish CWI corpus. A group of 40 native-speaking Spanish university students were tasked with identifying which words they believed to be complex within 3,887 academic texts. \citet{merejildo2021} conducted feature extraction on the identified complex words and found that word length and frequency were common markers of Spanish lexical complexity.}

\textcolor{black}{Apart from several researchers that have participated in CWI--2018 \cite{yiman-EtAl:2018:BEA} or that have later utilized the CWI--2018 dataset \cite{finnimore-etal-2019-strong, Aprosioetal2020}, little stand-alone research has been conducted on German LCP.} 

\subsubsection{\bf{Chinese}}
\citet{Lee&Yeung2018} created a SVM designed to identify Chinese complex words. Their monolingual LCP model was then further developed by \citet{yeung-lee-2018-personalized}. They tasked eight learners of Chinese to rank 600 Chinese words using a five-point likert scale. If the annotator assigned a complexity value of 1 to 3, then that word was labeled as complex. If, however, the word was assigned a complexity value of 4 or 5, then that word was labeled as being either challenging or non-complex respectively. Their SVM classifier was trained on a number of features parallel to \citet{Lee&Yeung2018}. These being, the target word's ranking in a Chinese proficiency test known as the Hanyu Shuiping Kaoshi \cite{Hanban2014}, along with word length, word frequency in the Chinese Wikipedia Corpus \cite{Lee&Yeung2018}, and character and word frequency in the Jinan Corpus of Learner Chinese \cite{wang-etal-2015-jinan}. They discovered that their logistic regression models outperformed their prior SVM \cite{Lee&Yeung2018}. They also found that their model was better at predicting the lexical complexities of their annotators with low Chinese L2 proficiency compared to those with high Chinese L2 proficiency. 

\subsubsection{\bf{Japanese}}
\citet{nishihara-kajiwara-2020-word} used a SVM to predict the lexical complexities of Japanese words. They created a new dataset that expanded upon the Japanese Education Vocabulary List (JEV). JEV contains 18,000 Japanese words divided into three levels of difficulty: easy, medium, or difficult. \citet{nishihara-kajiwara-2020-word} also rated the complexity of words from Japanese Wikipedia, the Tsukuba Web Corpus \cite{Tsukuba_corpus}, and the Corpus of Contemporary Written Japanese \cite{maekawa-etal-2010-design}. This increased the size of their dataset to 40,605 Japanese words. They trained a monolingual SVM to predict the level of complexity associated with each target word. To achieve this, they used a variety of features that were also used by prior English CWI systems, such as POS tags, character and word frequencies, and word embeddings. However, they discarded other popular features, such as word length, due to the topological and morphological differences between English and Japanese. Unlike English, Japanese “is composed of three types of characters: Hiragana, Katakana, and Kanji” \cite{nishihara-kajiwara-2020-word}. The characters Hiragana and Katakana are considered simple characters, whereas Kanji are ideographic and are therefore considered more difficult to interpret. As such, in Japanese, the proportion of complex to simple characters within a word is a good indicator of a word's complexity. \citet{nishihara-kajiwara-2020-word} concluded that the use of such language specific features was responsible for their model's good performance.

\subsubsection{\bf{Swedish}}
\citet{Smolenska_2018} experimented with a variety of models for binary CWI: SVM, RF, naïve bayes, gradient boosting, logistic regression, and stochastic descent models. These models were tested on one of two datasets consisting of Swedish words labeled with complexity ratings. The first dataset contained 4,305 Swedish words marked with labels from the Common European Reference Framework (CERF). These labels ranged from A1, elementary proficiency, to C2, advanced proficiency. The second dataset consisted of 4,238 manually extracted Swedish words from a variety of dictionaries and textbooks that were also labeled with CERF ratings. Whilst evaluating the quality of the two datasets, \citet{Smolenska_2018} discovered that the second dataset correlated better with the judgements of two human evaluators. Results showed that the RF model achieved the best performance on this dataset having been trained on a number of features, including morpho-syntactic, contextual, conceptual, and frequency based features.

\subsubsection{\bf{Multilingual LCP}}\label{multilingual}
\citet{sheang-2019-multilingual} saw the advantages of adopting a feature engineering approach as well as a CNN model for multilingual CWI. As a result, \citet{sheang-2019-multilingual} developed a semi-supervised CNN model trained on word embeddings and common CWI features, such as word frequency, word length, syllable and vowel count, term frequency, POS tags, syntactic dependency, and stop words. Being trained on the English, Spanish, and German datasets of CWI--2018 \cite{yiman-EtAl:2018:BEA} (Section \ref{CWI2018}), this multilingual model was found to outperform the best performing model of CWI--2018 \cite{yiman-EtAl:2018:BEA} on the Spanish and German datasets.

\citet{Aprosioetal2020} created a LCP system that caters for the native language of the user. As previously discussed within Section \ref{personalized_complexity}, an annotator's or user's native language influences their perception of lexical complexity through what is known as cross-linguistic influence. As such, their system was designed with the ability to identify the false friends as well as the cognates between the user's native language and the language of the annotated or inputted text. False friends are “those pairs of words in two different languages that are similar in form but semantically divergent” \cite{Aprosioetal2020}. Cognates, on the other hand, are those pairs of words with the same meaning and similar spelling in two or more languages. Their system firstly identified those words within the inputted text that may be considered cognate. It achieved this by taking into consideration three similarity metrics: XXDICE \cite{Brew1996WordPairEF}, Normalized Edit Distance \cite{Wagner_etal_1974}, and Jaro/Winkler \cite{winkler90}. Once potential cognates had been identified, their system used an SVM to classify which of these cognates may, in fact, be false friends. Their SVM was trained on the cosine similarity between the candidate words, and the cosine similarity between these words' synonyms. Those words which were found to be false friends were labeled as complex, whereas those words which were considered to be cognates and not false friends were labeled as non-complex. Thus, by taking the language of the user into consideration, \citet{Aprosioetal2020} created a LCP system that can recognize and exploit language-dependent features to improve its performance. \citet{Aprosioetal2020} is, therefore, another good example of personalized LCP. 

\subsubsection{\bf{Cross-Lingual LCP}}\label{cross-lingual_LCP}
\citet{finnimore-etal-2019-strong} continued working on CWI--2018's sub-task 1 \cite{yiman-EtAl:2018:BEA} (Section \ref{CWI2018}). They focused on the development of a cross-lingual CWI model with a particular focus on discovering which monolingual or multilingual CWI features would also perform well in a cross-lingual setting. They discarded previous features that they believed to be language-dependent, hence not transferable from one language to another. They state that the use of word-level n-grams is an example of such a language-dependent feature, since word-level n-grams denote the unique collocations of a particular language. Instead, \citet{finnimore-etal-2019-strong} experimented with a variety of features that they believed to be cross-lingual. These features being the number of syllables, tokens, and complex punctuation marks, along with the sentence length and character-level probabilities associated with a target word. They found that training their linear regression model on languages that belonged to the same language family as the target language improved its macro F1-score. However, the inclusion of languages unrelated to that of the target language had the opposite effect. Overall, their cross-lingual model achieved good performance. They go on to state that this is remarkable given its relatively simplistic set of features, thus proving the viability of cross-lingual LCP.

\citet{syspaper8} provide further evidence in favor of cross-lingual LCP. Their cross-lingual CWI system achieved the best F1-score in predicting the lexical complexities of an unseen language, being French. Consistent with other high performing CWI systems, it was an ensemble-based model that contained “a number of RFs as well as feed-forward neural networks with hard parameter sharing” \cite{syspaper8}. Their RFs were trained on a number of features, whereby they discovered that word length and frequency were good cross-lingual predictors of lexical complexity. 

\citet{zaharia2020crosslingual} experimented with several transformer-based models, such as Multilingual BERT (mBERT) \cite{pires-etal-2019-multilingual} and XLM-RoBERTa \cite{conneau-etal-2020-unsupervised}, for cross-lingual CWI. Both mBERT and XLM-RoBERTa are multilingual masked language models that are pretrained on numerous languages. mBERT is pretrained on “Wikipedia pages of 100 languages with a shared word piece vocabulary” \cite{pires-etal-2019-multilingual}. XLM-RoBERTa is also pretrained on 100 languages, yet with more data \cite{conneau-etal-2020-unsupervised}. \citet{zaharia2020crosslingual} tested these models performance on the WikiNews datasets provided by CWI--2018 \cite{yiman-EtAl:2018:BEA}. They found that XLM-RoBERTa was the best performing model. It achieved a higher F1-score than mBERT on the WikiNews datasets when tasked with predicting the lexical complexities of unseen German or French target words. These F1-scores being +0.02 and +0.04 respectively greater than that achieved by mBERT. They attributed XLM-RoBERTa's superior performance to its larger pretrained multilingual corpus \cite{zaharia2020crosslingual, conneau-etal-2020-unsupervised}.

\subsubsection{Is transfer learning possible for predicting lexical complexity across multiple languages?}\label{RQ6}

\textcolor{black}{The studies detailed in Section \ref{cross-lingual_LCP} provide evidence in favor of transfer learning for cross-lingual LCP. Numerous features, for instance, number of syllables, tokens, complex punctuation marks, and sentence length, have been proven to work well when trained on one language and then used to predict lexical complexities in another \cite{syspaper8, finnimore-etal-2019-strong}. Models, such as mBERT and XLM-RoBERTa have also been shown to achieve good performances for cross-lingual LCP \cite{zaharia2020crosslingual}. With the availability of LCP datasets in high-resource languages (Section \ref{resources}) and with LCP research gaining traction in languages other than English (Section \ref{other_languages}), we suspect cross-lingual LCP will become \textcolor{black}{increasingly} popular.}

\section{Summary}\label{summary}

This paper has presented an overview of LCP research with a specific focus on research conducted on English. It has defined what is meant by "\textit{complexity}" within LCP and has described types of computational modelling applied to its prediction, such as comparative, binary, continuous, and personalized complexity (Sections \ref{complexity} to \ref{types_LCP}). It has provided the evaluation metrics used to evaluate LCP performance and has discussed the international shared-tasks that have inspired the creation of numerous LCP systems (Sections\ref{evaluation_metrics} to \ref{shared-tasks}): CWI--2016 \cite{paetzold-specia:2016:SemEval1}, CWI--2018 \cite{yiman-EtAl:2018:BEA}, and LCP--2021 \cite{semeval-2021}. It has explained the architecture, development, and evolution of these LCP systems, ranging from feature engineering approaches, neural networks to the most recent state-of-the-art transformer-based models whilst discussing relevant research questions within the field (Section \ref{system_reports}). It has presented various use cases and applications of LCP, including for other NLP-related tasks such as machine translation (Section \ref{MT}) and author identification (Section \ref{AI}). It has collected and summarized English datasets (Section \ref{English_datasets}) and has also briefly presented work on languages other than English (Section \ref{other_languages}).

\subsection{Opportunities and Challenges}

\textcolor{black}{There now exists an unprecedented demand for LCP research. With distance learning becoming ever more popular and with LCP being a precursor within other NLP-related tasks, the future for LCP research would appear to be promising. LCP--2021 \cite{semeval-2021} has shown the superiority of transformer-based models for LCP, especially when a diverse set of transformers are used to form an ensemble-based model \cite{semeval2021_task1_paper_148, semeval2021_task1_paper_74}. CWI--2018 along with others \cite{finnimore-etal-2019-strong, syspaper8, zaharia2020crosslingual}, have proven that cross-lingual LCP is viable. LCP is now being conducted for languages other than English \cite{Lee&Yeung2018, yiman-EtAl:2018:BEA,  Smolenska_2018,  Zambrano2020OverviewOA, Tack2021, nishihara-kajiwara-2020-word} . As such, we expect to see ensemble-based models with a diverse set of transformers being used for multi-lingual and cross-lingual LCP. Furthermore, personalized LCP calls for the development of LCP systems with the ability to predict the complexity assignments made by the individual or specific target demographic, rather than belonging to a generalized population \cite{Zengetal_2005, tack_etal_2016, lee-yeung-2018-personalizing}. We expect such personalized LCP systems to become popular as their datasets are likely to contain more consistent complexity ratings due to there being less disagreement among their annotators. Research questions investigating such areas as the effect of including context on LCP performance as well as the advantages of complexity prediction of multi-word expressions, are other avenues of LCP research that have likewise proven to aid LCP \cite{goodingetal2020, semeval2021_task1_paper_148, semeval2021_task1_paper_148}. We therefore also believe that context and MWEs will continue to be taken into consideration by future LCP systems.}

\textcolor{black}{Future LCP research, however, is not without its challenges. A current lack of available data may have already lead to some cases of overfitting with models being unable to generalize their predictions across multiple domains or target populations. In addition, dataset quality has previously been put into question, whereby the use of a small pool of annotators, an irregular train/test split, or high levels of inter-annotator disagreement may have lead to unreliable complexity labels \cite{zampieri-EtAl:2017:NLPTEA}. To overcome these challenges, we stress the importance of further research into continuous and personalized complexity prediction that takes inconsideration context and MWEs, along with the implementation of transfer-learning models for under-resourced languages.}

\section*{Acknowledgments}

The authors would like to thank Richard Evans for the valuable suggestions and feedback provided. We further thank the anonymous ACM CSUR reviewers for their insightful feedback.  


\bibliographystyle{ACM-Reference-Format}
\bibliography{CWI,Matt,Thesis,bibliography_Matt,bibliographyEvans,bibliographymarcos,semeval_2021_participants}

\newpage

\section*{Appendices} \label{appendices}

\begin{table*}[!ht]
\centering
\scalebox{0.76}{
  \begin{tabular}{lp{5cm}p{10cm}c}
\hline
  \bf Team & \bf Classifiers & \bf Features & \bf Paper \\ \hline
  
  \bf AI-KU & SVM & word embeddings of the target and surrounding words & \cite{kuru:2016:SemEval} \\
  
  \bf  Amrita-CEN & SVM & word embeddings and various semantic and morphological features & \cite{sp-kumar-kp:2016:SemEval} \\
    
\bf     BHASHA & SVM, Decision Tree & lexical and morphological features & \cite{choubey-pateria:2016:SemEval} \\
  
 \bf ClacEDLK & Random Forests & semantic, morphological, and psycholinguistic features & \cite{davoodi-kosseim:2016:SemEval}\\
  
 \bf  CoastalCPH & Neural Network, Logistic Regression & word frequencies and word embeddings & \cite{bingel-schluter-martinezalonso:2016:SemEval} \\
  
\bf    HMC & Decision Tree and Random Forest & lexical, semantic, syntactic and psycholinguistic features & \cite{quijada-medero:2016:SemEval} \\
    
  \bf  IIIT & Nearest Centroid & semantic and morphological features & \cite{palakurthi-mamidi:2016:SemEval} \\
    
    \bf JUNLP & Random Forest, naïve Bayes & semantic, lexicon-based, morphological and syntactic features & \cite{mukherjee-EtAl:2016:SemEval} \\
     
\bf     LTG & Decision Tree & n-grams and word length & \cite{malmasi-dras-zampieri:2016:SemEval} \\

  \bf    MACSAAR & Random Forest, SVM & Zipfian frequency distribution, word length & \cite{zampieri-tan-vangenabith:2016:SemEval} \\
    
\bf  MAZA & Meta-classifier & n-grams, word probability, word length & \cite{malmasi-zampieri:2016:SemEval} \\
  
 \bf   Melbourne & Weighted Random Forests & lexical and semantic features & \cite{brooke-uitdenbogerd-baldwin:2016:SemEval} \\
  
 \bf   PLUJAGH & Threshold-based methods & features extracted from Simple Wikipedia & \cite{wrobel:2016:SemEval} \\
  
 \bf   Pomona & Threshold-based methods & word frequencies & \cite{kauchak:2016:SemEval} \\
  
 \bf   Sensible & Ensemble Recurrent Neural Networks & word embeddings & \cite{nat:2016:SemEval} \\
  
 \bf SV000gg & System voting with threshold & morphological, lexical, and semantic features & \cite{paetzold-specia:2016:SemEval2} \\
  
\bf  TALN & Random Forest & lexical, morphological, semantic, and syntactic features & \cite{ronzano-EtAl:2016:SemEval} \\
  
\bf    USAAR & Bayesian Ridge classifiers & hand-crafted word sense entropy metric and language model perplexity & \cite{martinezmartinez-tan:2016:SemEval} \\
  
\bf  UWB & Maximum Entropy & word occurrence counts on Wikipedia documents & \cite{konkol:2016:SemEval} \\

  \hline
  \end{tabular}
}
\caption{Systems submitted to the CWI--2016 in alphabetical order as summarized by \cite{shardlow2021predicting}.}
\label{table_CWI2016}
\end{table*}


\begin{table*}[!ht]
\centering
\scalebox{0.76}{
  \begin{tabular}{lp{5cm}p{9.8cm}c}
\hline
  \bf Team & \bf Classifiers & \bf Features & \bf Paper \\ \hline

\bf   Camb & Adaboost & N-grams, WordNet features, POS tags, dependency parsing relations, psycholinguistic features. & \cite{syspaper7} \\

\bf   CFILT\_IITB & Voting ensemble & Word length, syllable counts, vowel counts, WordNet-based features. & \cite{syspaper10} \\

\bf   hu-berlin & naïve Bayes & Character n-grams & \cite{syspaper3} \\

\bf   ITEC & LSTM & Word length, word and character embeddings, frequency count, psycholinguistics features. & \cite{syspaper6}\\

\bf   LaSTUS/TALN & SVM, Random Forest & Word length, word embeddings, semantic and contextual features.  & \cite{syspaper9} \\

\bf   NILC &  XGBoost & N-grams, word length, number of syllables, WordNet-based features. & \cite{syspaper4} \\

\bf   NLP-CIC & Tree Ensembles and CNNs & Word frequency, syntactic and lexical features, psycholinguistic features, and word embeddings.  &  \cite{syspaper11} \\

\bf   SB@GU & Extra Trees & Word length, number of syllables, n-grams, frequency distribution. & \cite{syspaper2} \\

\bf   TMU & Random Forest & Word length, word frequency, probability features derived from corpora. &\cite{syspaper5} \\

\bf  UnibucKernel & Kernel-based learning with SVMs. & Character n-grams, semantic features, and word embeddings. & \cite{syspaper1} \\

  \hline
  \end{tabular}
}
\caption{Systems submitted to the CWI--2018 English binary classification single word track in alphabetical order as summarized by \cite{shardlow2021predicting}.}
\label{tab:approaches2018}
\end{table*}


\begin{table*}[!ht]
\centering
\scalebox{0.76}{
  \begin{tabular}{lp{4.8cm}p{8.8cm}c}
\hline
  \bf Team & \bf Classifiers & \bf Features & \bf Paper \\ \hline
  
  Alejandro Mosquera  & Gradient Boosted Regression & Length, Frequency, Semantic, Sentence & \cite{semeval2021_task1_paper_62}\\
  
Andi & Ridge Regression, Gradient Boosted Regression  & Psycholinguistic, Glove, Word2Vec, ConceptNet NumberBatch, BERT, RoBERTa, ELECTRA, ALBERT, DeBERTa & \cite{semeval2021_task1_paper_147} \\

Archer & Random Forest Regression, Gradient Boosted Regression & Length, Frequency, Psycholinguistic, Scrabble Score, Word Inclusion, Semantic  & \cite{semeval2021_task1_paper_186}\\

BigGreen  & Gradient Boosted Regression, BERT  & Length, Semantic, Glove, Elmo, InferSent, Phonetic, Frequency, POS & \cite{semeval2021_task1_paper_155} \\

C3SL   & Multi-layer Perceptron & Sent2Vec & \cite{semeval2021_task1_paper_163} \\

Cambridge  & BERT, Random Forest Regression  & Frequency, Syntactic, Length & \cite{semeval2021_task1_paper_77}\\

CLULEX   & Decision Tree & Frequency, POS, Named Entities, Word Inclusion, Sentence, Bert & \cite{semeval2021_task1_paper_120} \\

CompNA   & Decision Tree Ensemble & Length, Semantic, Glove, Word Inclusion, & \cite{semeval2021_task1_paper_64}\\

CS-UM6P   & BERT, RoBERTa  & Token and Context Encoded & \cite{semeval2021_task1_paper_76}\\

CSECU-DSG    & BERT, RoBERTa & Token and Context Encoded & \cite{semeval2021_task1_paper_117}\\

DeepBlueAI   & BERT, ALBERT, RoBERTa, ERNIE & Token and Context Encoded& \cite{semeval2021_task1_paper_74} \\

IA PUCP    & Gradient Boosted Regression & Sentence, POS, N-gram Frequency, RoBERTa, XLNet, BERT & \cite{semeval2021_task1_paper_195} \\

IITK@LCP   & Linear Regression, Support Vector Machine  & ELECTRA + Glove& \cite{semeval2021_task1_paper_57} \\

JCT   & Gradient Boosted Regression & POS, Frequency, BERT, Cluster Features & \cite{semeval2021_task1_paper_156}\\

JUST-BLUE   & Average of Weighted Bert and Roberta  & Token Encoded and Context Encoded & \cite{semeval2021_task1_paper_148}\\

Katildakat   & Linear Regression, Multi-layer Perceptron & BERT, Length, BERT-score, Frequency, Semantic, & \cite{semeval2021_task1_paper_198} \\

LAST   & Gradient Boosted Regression & Frequency, Psycholinguistic, Sentence, Bigram Association & \cite{semeval2021_task1_paper_70} \\

LCP-RIT    & Random Forest Regressor & Length, Frequency, Character N-Grams, Psycholinguistic, POS& \cite{LCP-RIT} \\

LRL\_NC   & Random Forest Regressor  & Frequency, Semantic, Language Model, Psycholinguistic, Word Inclusion & \cite{semeval2021_task1_paper_56}\\

Hub   & RoBERTa, Inception & TF-IDF, Context Encoded  & \cite{semeval2021_task1_paper_86}\\

Manchester Metropolitan  & CNN  & Frequency, Psycholinguistic, Length, Embeddings & \cite{semeval2021_task1_paper_91} \\

OCHADAI-KYOTO   & BERT, RoBERTa & Token and Context Encoded & \cite{semeval2021_task1_paper_67} \\

PolyU CBS-Comp   & Gradient Boosted Regression & Frequency, Length, Capitalisation, POS, Embeddings, BERT, GPT-2 & \cite{semeval2021_task1_paper_68}\\

RG PA   & RoBERTa & Context Encoded& \cite{semeval2021_task1_paper_107}\\

RS\_GV  & Feed-Forward Neural Network  & GLoVE, ELMo, BERT, Flair, Readability, Length, Frequency, Semantic, Psycholinguistic, Morphological, Word Inclusion, Named Entity & \cite{semeval2021_task1_paper_139}\\

Stanford MLab    & Gradient Boosted Regression & Glove, Length, POS, Named Entity  & \cite{semeval2021_task1_paper_180}\\

TUDA-CCL    & Gradient Boosted Regression & Linguistic, Semantic, Embeddings, Psycholinguistic, Frequencies, Word Inclusion & \cite{semeval2021_task1_paper_150}\\

UNBNLP   & Neural Network, Support Vector Machine & Length, Frequency, Character-Level-Encoder, BERT & \cite{semeval2021_task1_paper_142}\\

UPB   & BERT, RoBERTa, Linear Regression & Transformers, Word Embeddings, Character Wmbeddings, Length, Psycholinguistic  & \cite{semeval2021_task1_paper_96}\\

UTFPR   & Support Vector Machine & Frequency, Length, Semantic, Bert Embedding & \cite{semeval2021_task1_paper_99}\\

  \hline
  \end{tabular}
}
\caption{Systems submitted to LCP--2021 in alphabetical order as shown in \cite{semeval-2021}.}
\label{table_LCP2021}
\end{table*}

\begin{table*}[!ht]
\centering
\scalebox{0.76} {
  \begin{tabular}{p{2.5cm}p{3cm}p{4cm}p{4cm}p{3cm}c}
\hline
  \bf Dataset & \bf Complexity & \bf Size & \bf Annotators  & \bf Noteworthy Comments & \bf Paper  \\ \hline
  
\bf   LS--2012 & Comparative & 201 complex words each with several candidate simplifications. & Native English speakers provided simplifications and 4 L2 learners ranked these simplifications based on their complexity.  & Each Complex word is shown in 10 different contexts.  & \cite{specia2012} \\

\bf   CW Corpus & Binary - Comparative & 731 complex words and their equivalent simplification. & Complex words gained via Wikipedia edit history, editor comments, and a series of simplification checks. & Complex words are provided with context. & \cite{CWcorpus} \\ 

\bf   \citet{Horn2014} & Comparative & 500 complex words each with 50 candidate simplifications. & 50 annotators from the US. &  Data was acquired from the sentence-aligned Wikipedia corpus. Complex words are also provided with context. & \cite{Horn2014} \\

\bf   CWI--2016 & Binary & 35,958 tokens with 232,481 instances, 3,854 of these tokens were labeled as complex. & 400 non-native English speakers from a mix of international and educational backgrounds with varying levels of English proficiency. & Training set included 2,237 target words in 200 sentences, whereas the test set included 88,221 target words in 9,000 sentences. & \cite{paetzold-specia:2016:SemEval1} \\ 

\bf   CWI--2018 & Binary - Continuous & 34,789 English, 7,905 German, 17,605 Spanish, and 2,251 French words. Out of these, 14,428, 3,272, 7015, and 657 were complex respectively. & A mix of native and non-native speaking annotators for a variety of international backgrounds. & Datasets were also divided on source: News, WikiNews, and Wikipedia.& \cite{yiman-EtAl:2018:BEA} \\

\bf   Word Complexity Lexicon & Continuous & 15,000 words labeled with varying degrees of complexity. & 11 non-native yet fluent English speakers. &  Used a dictomous six-point likert scale for annotation: very easy (very hard), moderately easy (moderately hard), and easy (hard). & \cite{maddela2018word} \\ 

\bf   Personalized LS Dataset & Personalized - Continuous  & 12,000 words labeled with varying degrees of complexity. & 15 learners of English, who were native Japanese speakers.  & Used a five-point likert scale for annotation. Scores 1-4 were deemed as complex, whereas 5 was considered non-complex & \cite{lee-yeung-2018-personalizing} \\

\bf   CompLex Dataset & Continuous & 10,800 words and MWEs labeled with varying degrees of complexity. & Annotators were crowd sourced from the US, UK, and Australia. A median of 7 annotators labeled each word. & Used a five-point likert scale for annotation. Scores were directly converted into continuous complex values. Words in context were taken from the Bible, biomedical articles, and europarl. & \cite{shardlow-etal-2020-complex} \\

  \hline
  \end{tabular}
}
\caption{Datasets used for English complexity prediction research arranged in chronological order.}
\label{tab:datasets}
\end{table*}

\end{document}